\documentclass[10pt]{article}

\usepackage[letterpaper, left=1in, top=1in, right=1in, bottom=1in, verbose, ignoremp]{geometry}

\usepackage{url}\RequirePackage[colorlinks,citecolor=blue, linkcolor=blue,urlcolor = blue]{hyperref}
\usepackage{latexsym,amssymb,amsmath,amsfonts,graphicx,color,fancyvrb,amsthm,enumerate,mathrsfs, caption, subcaption}
\usepackage[authoryear,round]{natbib}
\usepackage[dvipsnames]{xcolor}
\usepackage{xy}\xyoption{all} \xyoption{poly} \xyoption{knot}
\usepackage{float}
\usepackage{bm}
\usepackage{bbm}
\usepackage{multicol,multirow}
\usepackage{array}
\usepackage{relsize}
\usepackage{chngcntr}
\usepackage{etoolbox}
\usepackage{tikz}
\usetikzlibrary{decorations.pathreplacing,calc}
\usetikzlibrary{shapes,backgrounds}
\usetikzlibrary{patterns}
\usepackage{amsopn}
\usepackage{calc}
\usepackage{hyperref}
\usepackage{mathtools}
\usepackage{adjustbox}
\usepackage{textpos}
\usepackage{tabularx}
\usepackage{algorithm}
\usepackage{algpseudocode}

\newtheorem{theorem}{Theorem}[]
\newtheorem*{theorem*}{Theorem}

\newtheorem*{claim*}{Claim}
\theoremstyle{definition}
\newtheorem{definition}[theorem]{Definition}
\newtheorem*{definition*}{Definition}
\theoremstyle{AppDefinition}

\theoremstyle{AppClaim}

\theoremstyle{remark}

\newtheorem{example}[theorem]{Example}
\newtheorem*{example*}{Example}

\makeatletter
\newcommand*{\op}{%
  \DOTSB
  \mathop{\vphantom{\bigoplus}\mathpalette\matt@op\relax}%
  \slimits@
}
\newcommand\matt@op[2]{%
  \vcenter{\m@th\hbox{\resizebox{\widthof{$#1\bigoplus$}}{!}{$\boxplus$}}}%
}
\makeatother

\newcommand{\br}{\vspace{5mm}}

\makeatletter
\def\@biblabel#1{}
\makeatother

\makeatletter
\patchcmd{\NAT@citex}
  {\@citea\NAT@hyper@{%
     \NAT@nmfmt{\NAT@nm}%
     \hyper@natlinkbreak{\NAT@aysep\NAT@spacechar}{\@citeb\@extra@b@citeb}%
     \NAT@date}}
  {\@citea\NAT@nmfmt{\NAT@nm}%
   \NAT@aysep\NAT@spacechar\NAT@hyper@{\NAT@date}}{}{}

\patchcmd{\NAT@citex}
  {\@citea\NAT@hyper@{%
     \NAT@nmfmt{\NAT@nm}%
     \hyper@natlinkbreak{\NAT@spacechar\NAT@@open\if*#1*\else#1\NAT@spacechar\fi}%
       {\@citeb\@extra@b@citeb}%
     \NAT@date}}
  {\@citea\NAT@nmfmt{\NAT@nm}%
   \NAT@spacechar\NAT@@open\if*#1*\else#1\NAT@spacechar\fi\NAT@hyper@{\NAT@date}}
  {}{}

\DeclarePairedDelimiter{\abs}{\lvert}{\rvert}

\makeatother

\algblock{Input}{EndInput}
\algnotext{EndInput}
\algblock{Output}{EndOutput}
\algnotext{EndOutput}

\begin{document}
\def\spacingset#1{\renewcommand{\baselinestretch}%
{#1}\small\normalsize} \spacingset{1}
%\spacingset{1.45} % DON'T change the spacing!
\begin{flushleft}
{\Large{\textbf{Rewiring Networks for Graph Neural Network Training\\ Using Discrete Geometry}}}
\newline
\\
Jakub Bober$^{1,2}$, Anthea Monod$^{1,\dagger}$, Emil Saucan$^{3}$, and Kevin N.~Webster$^{1,4}$
\\
\bigskip
\bf{1} Department of Mathematics, Imperial College London, UK
\\
\bf{2} Department of Computing, Imperial College London, UK
\\
\bf{3} Department of Applied Mathematics, ORT Braude College of Engineering, Karmiel, Israel
\\
\bf{4} FeedForward Ltd., London, UK
\bigskip

$\dagger$ Corresponding e-mail: a.monod@imperial.ac.uk
\end{flushleft}

%%%%%%%%%%%%%%%%%%%%%%%%%%%%%%%%%%%%%%%%%%%%%%%%%%%

\section*{Abstract}
Information over-squashing is a phenomenon of inefficient information propagation between distant nodes on networks.  It is an important problem that is known to significantly impact the training of graph neural networks (GNNs), as the receptive field of a node grows exponentially.  To mitigate this problem, a preprocessing procedure known as rewiring is often applied to the input network.  In this paper, we investigate the use of discrete analogues of classical geometric notions of curvature to model information flow on networks and rewire them.  We show that these classical notions achieve state-of-the-art performance in GNN training accuracy on a variety of real-world network datasets.  Moreover, compared to the current state-of-the-art, these classical notions exhibit a clear advantage in computational runtime by several orders of magnitude. 

\paragraph{Keywords:} Discrete curvature, geometric deep learning, graph neural networks, graph rewiring, information over-squashing.

\section{Introduction}
\label{sec:into}

The abundance of data availability has resulted in the occurrence of data captured by structures beyond vectors living in Euclidean space.  Much fundamental information is encoded in data that exhibit more complex structures, some with a distinct geometric characterization---such as networks.  The driving premise underlying  \emph{geometric deep learning} is that this geometry encompasses information that is crucial to take into consideration when developing machine learning techniques (specifically, deep learning) to handle these data \citep{geometric-learning}.  Thus, the inherent, non-Euclidean geometry of the data structures as well as the space they live in are fundamental aspects to understand and build into deep learning architectures.

%\textit{Geometric deep learning} is a notion introduced in a 2017 paper by Michael Bronstein and co-authors \cite{geometric-learning}. It is an umbrella term that aims to unify certain branches of machine learning using geometric approach via the study of \textit{symmetry}. It has been named "The Erlangen Programme of ML" at the ICLR 2021 talk, when the paper \cite{geometric-learning} was first introduced \footnote{\url{https://iclr.cc/virtual/2021/invited-talk/3717}}.

%Geometric deep learning defines the task of performing Machine Learning, specifically Deep Learning, on non-euclidean data. It means that the data is not represented in a Euclidean domain, but instead in a space with a different notion of distance. A prime example of this is data represented by mathematical graphs, as the distance between graph nodes is not Euclidean, but rather expressed by graph edges. As the order of nodes in a graph can be arbitrary and is of no semantic significance (graphs are \textit{permutation-invariant}), this motivates the study of symmetry via geometric methods. The topic of Deep Learning on graph data is recently gaining a lot of interest in the research community. Specifically, there is a high output of brand-new research on Graph Neural Networks (GNNs) \cite{gnn, bottleneck-bronstein, sampling, cgnn}.

In this paper, we consider network data and an important problem associated with training graph neural networks (GNNs).  Specifically, we study the problem of \emph{information over-squashing} \citep{bottleneck,bottleneck-bronstein}, which amounts to inefficient information propagation between distant nodes on a graph.  This phenomenon is especially significant in tree-like graphs, where multiple nodes lead to a single node---namely, the ``bottleneck.''  A common approach to mitigate this problem is to perform graph rewiring on the input network data by adding or suppressing edges in the network in order to alleviate such bottlenecks and increase the efficiency of information flow over a network.  Recent pioneering work by \cite{bottleneck-bronstein} models information flow on a network using notions of discrete curvature and uses this network curvature information to perform graph rewiring prior to training GNNs, yielding the current state-of-the-art for GNN training in the presence of bottlenecks.  In particular, a novel discrete curvature, the \emph{balanced Forman curvature}, was introduced and utilized to identify bottlenecks and rewire graphs prior to training for increased efficiency in information propagation over networks \citep{bottleneck-bronstein}.

The discretization of classical notions of smooth geometry has been actively studied in recent decades \citep[see for instance,][]{najman-romon}, resulting in various definitions of discrete curvature.  An important motivation behind such discretizations are for the application of geometric methods to statistics and machine learning tasks for data exhibiting discrete geometric structure, such as network learning by sampling \citep[e.g.,][]{barkanass2020geometric,sigbeku2021curved}.  In this work, we return to these fundamental principles and study the alleviation of information over-squashing by graph rewiring following the procedure used by \cite{bottleneck-bronstein}.  We systematically test and compare the performance of several classical discrete curvature notions against the recently proposed balanced Forman curvature on several benchmarking datasets, and find that these classical discrete curvature notions are able to achieve state-of-the-art performance in terms of accuracy for GNN training.  Moreover, the computation of these classical discrete curvatures is much more efficient and runs several orders of magnitude faster than the state-of-the-art.\\

The remainder of this paper is organized as follows. Section \ref{sec:gnn} discusses in further detail the GNNs and the problem of information over-squashing, and briefly surveys various approaches to reducing over-squashing.  In Section \ref{sec:math}, we then switch to discussing mathematical details on discrete curvature and formally present the notions studied in this paper; we also present the balanced Forman curvature recently proposed by \cite{bottleneck-bronstein}.  We also overview the procedure to identifying network bottlenecks and performing graph rewiring adapted by \cite{bottleneck-bronstein}, which use discretizations of smooth curvature concepts.  This will be the same procedure that we will implement with the various classical discrete curvature notions.  In Section \ref{sec:methods}, we describe the data we study and our experimental design and setup.  In Section \ref{sec:results}, we demonstrate the method on a wide variety of benchmarking datasets and present the accuracy and computational runtime results.  Finally, we close in Section \ref{sec:end} with a discussion and some proposals for future research.

%%%%%%%%%%%%%%%%%%%%%%%%%%%%%%%%%%%%%%%%%%%%%%%%%%%%%%%%%%%%%%%%%%%%
\section{Graph Neural Networks and Information Over-Squashing}
\label{sec:gnn}

Neural networks are systems of algorithms that aim to identify underlying relationships in data in a manner similar to how biological neural networks in brains function.  They consist of collections of artificial ``neurons'' and ``synapses'' that are typically organized into layers.  Among deep neural networks, a specific class are adapted specifically to  handling graph or network data---collections of vertices connected by edges that mathematically describe a dependency structure----which is the focus of this paper.

The main difference between the traditional deep neural networks and GNNs lies in the functioning of the \textit{message passing} algorithm \citep{message-passing}.  Briefly, in message passing, at each layer and for each node, features from the neighboring nodes are aggregated before updating the features of the target node. This is the mechanism by which the network captures the information from the graph structure of the data. %The simplest type of such network is the Message Passing Neural Network (MPNN) \cite{message-passing}. More complex architectures can be built upon the MPNN framework, such as Graph Convolutional Network (GCN) \cite{gcn} or Curvature Graph Neural Network (CGNN) \cite{cgnn}.

\subsection{Information Over-Squashing}
% \label{section:over-squashing}

Training GNNs presents new issues in comparison to standard neural network training due to the discrete geometric structure of network data.  A particularly important challenge that has recently gained much research interest is that of \textit{information over-squashing}, also known as the \emph{bottleneck problem} \citep[e.g.,][]{bottleneck, bottleneck-bronstein}.

In over-squashing, the principle concern is that the influence of certain node features (which may be important) may be too small and eventually have minimal or no impact on features of distant nodes on the network when performing message passing over the GNN.  This is particularly problematic in the context of network data, since the receptive field of a graph node is known to grow exponentially \citep{bottleneck}.  
%(note that the distance between two nodes is defined to be the number of edges on the shortest path between these nodes).

\begin{figure}
\centering
\includegraphics[width=0.9\linewidth]{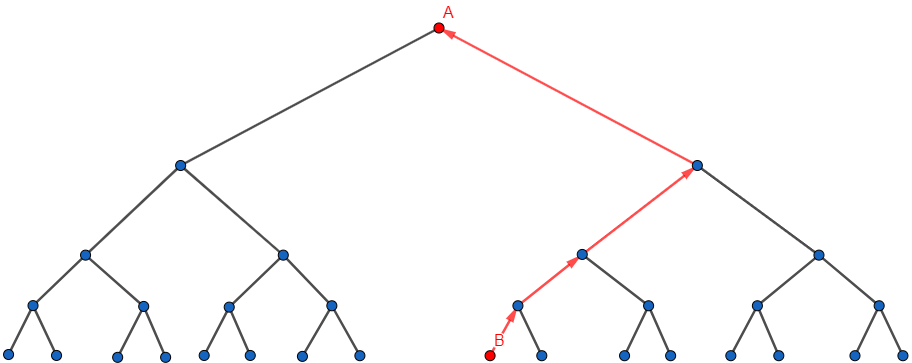}
\caption{The receptive field of the root node in the binary tree grows exponentially. When the information from node $B$ reaches node $A$, it will be ``squashed'' together with information from all of the other nodes in the righthand subtree.}
\label{fig:binary-tree}
\end{figure}

\begin{example}
In a binary tree, let the $n$-jump neighborhood of a node $i$ be the set of nodes in the graph that contains only the nodes that have the shortest path of at most $n$ to the node $i$.  Then the receptive field of the root doubles with every jump: there are twice as many nodes in the $k+1$-jump neighborhood than in the $k$-jump neighborhood for any integer $k>0$.  See Figure~\ref{fig:binary-tree}.
\end{example}

Thus, over-squashing is a crucial issue to take into consideration, especially when the long-range dependencies in the data are important for the learning task when training on graph data.  It is primarily caused by the poor propagation of long-distance information by some specific edges in the graph. As an illustration, consider two \textit{components} that are cliques (or two densely connected, clique-like graphs) connected only by a single edge, illustrated in Figure~\ref{fig:rewiring1}.  When propagating information from a node in a source component to a node in the target component, over-squashing is likely to happen as the information is crowded or ``squashed'' together with all other node features from the source component. This happens on the edge connecting the two components which, here, is the main source of over-squashing in the graph and called a \textit{bottleneck}.

\subsection{Mitigating Over-Squashing}

\begin{figure}
\centering
\begin{subfigure}[t]{0.49\textwidth}
\centering
\includegraphics[width=0.9\linewidth]{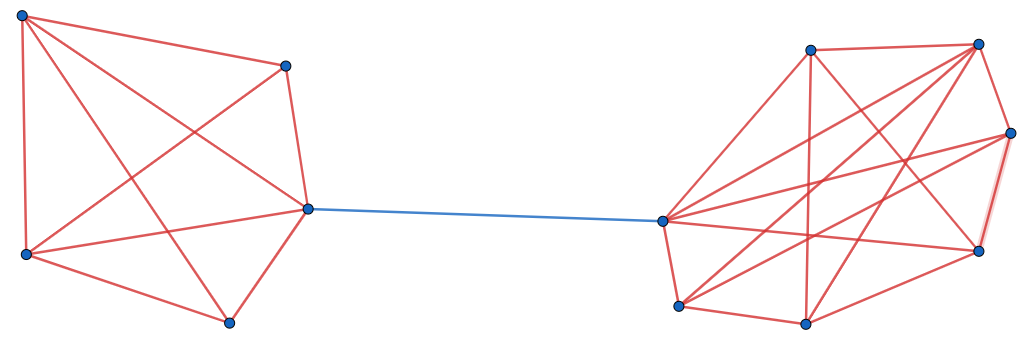}
\caption{}
\label{fig:rewiring1}
\end{subfigure}
\begin{subfigure}[t]{0.49\textwidth}
\centering
\includegraphics[width=0.9\linewidth]{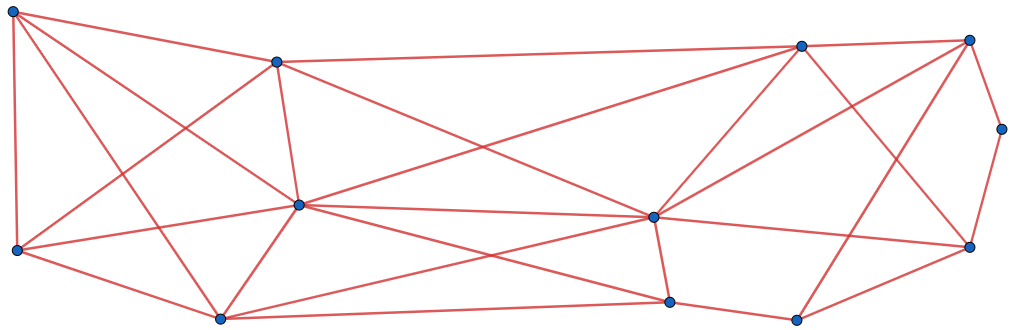}
\caption{}
\label{fig:rewiring2}
\end{subfigure}
\caption{Graph rewiring reduces over-squashing. (\subref{fig:rewiring1}): A graph with a bottleneck (blue edge). (\subref{fig:rewiring2}): A possible rewiring that alleviates the bottleneck.}
\label{fig:rewiring}
\end{figure}

Bottlenecks may be alleviated to reduce over-squashing via \textit{graph rewiring}, which adds or suppresses edges in the graph to obtain a new graph with the same nodes and node features, but a different set of edges. The goal of the rewiring is to better support the bottleneck and give alternative routes of access between components which improves the information flow between components and reduces the risk that features become crowded out (over-squashed). Edges that have little impact on the flow of information in the graph can be deleted to control the size of the graph. An example of a rewired graph with an alleviated bottleneck is shown in Figure~\ref{fig:rewiring2}.

Several approaches to bottleneck alleviation have been proposed in the recent literature.  For example, \cite{digl} propose graph diffusion convolution (GDC) as a graph rewiring approach using a discretization of the gas diffusion equation to model the propagation of information on a network.  However, this method fails to capture long-range dependencies on networks \citep{bottleneck-bronstein}.  There also exist other bottleneck alleviation methods that do not entail rewiring.  For example, much in the spirit of the work of this paper, \cite{cgnn} propose a curvature GNN (CGNN) which, instead of adding or deleting edges, assigns specific weights to graph edges as a measure of how much information flows over this edge where the weights are determined by discrete curvature.  The work of \cite{bottleneck-bronstein} uses a hybrid of these two methods where graph rewiring is performed driven by a newly proposed discrete curvature.

%Just as SDRF discretises the idea of Ricci curvature and Ricci flow of manifolds, in \cite{digl}, a \textit{Graph Diffusion Convolution} (GDC) is proposed as the discretisation of the idea of gas diffusion. The adjacency matrix of the graph is \textit{smoothed out}. In \cite{bottleneck-bronstein}, it is argued that the Cheeger constant cannot be improved arbitrarily well using this rewiring, and that this method can fail to improve the meaning of the very long-range dependencies. It is also presented that the final accuracy using SDRF is better for the majority of datasets. Moreover, the diffusion-based rewiring adds hundreds of times more edges than SDRF and does not compensate it with deleting other edges \cite{bottleneck-bronstein}, so the structure of the graph is drastically changed, which is not a desired property.

%Another attempt to reduce information over-squashing, introduced as a Curvature Graph Neural Network \cite{cgnn} (CGNN) does not use graph rewiring at all. Instead, it assigns specific weights to graph edges. The weights are specified by (1D Forman or Ollivier) discrete curvature (made positive and normalised, to make the weights more meaningful). The effect is that the edges that suffer from information over-squashing are given higher weights to compensate for the squash. The main advantage of such approach is that the graph structure is not changed at all. CGNN is reported to achieve state-of-the-art results for the benchmarking datasets. More information can be found in \cite{cgnn}, where this architecture is proposed.

While graph rewiring effectively alleviates the over-squashing problem, it does present limitations.  First, on some types of data, the rewiring approach may not be applicable at all; an example is in chemical data where the graphs represent molecules and adding or deleting edges changes the molecule under study entirely.  Second, rewiring alters the structure (topology) of the graph and changes the information that may be captured from the graph connectivity, which can negatively impact feature recognition \citep{bottleneck}.  In this case, there is a trade-off between reducing bottlenecks and changing the graph topology to consider.  It is therefore important to obtain a measure of severity of the bottleneck---the \emph{bottleneckness} of the graph---and and use it as a guide when performing graph rewiring.

\subsection{Quantifying Over-Squashing}

In this paper, we use the Jacobian as a measure of the bottleneckness of a graph.  Consider a graph with $n$ nodes; take two nodes $i, j$ that are at a distance $d$ from each other, where $d$ is the number of edges on the shortest path between these nodes.  To quantify over-squashing, we need to measure how much of an impact $x_i$ the feature vector of $i$ has on the feature vector of $j$ after $d$-many forward passes (i.e., message passing is performed $d$ times); denote this impact by $h_j^{(d)}(x_i)$.  Then the Jacobian
\begin{equation}
    \label{eq:jacobian}
    \abs*{\frac{\partial h_j^{(d)}}{\partial x_i}(x_i)} \quad \forall\ 0\leq i \leq n - 1, 0 \leq j \leq n - 1
\end{equation}
for $d$-distance dependencies quantifies the over-squashing in a graph \citep{bottleneck-bronstein}.

%In other words, the point is to see how the change in $x_i$ changes $h_j^{(d)}$. This can be captured by the partial differential $\frac{\partial h_j^{(d)}}{\partial x_i}$. 

\cite{bottleneck-bronstein} show that the bounds on the elements of the Jacobian of $d$-distance dependencies are proportional to the respective $d$th powers of the normalized augmented adjacency matrix, i.e.,
\begin{equation}
\label{eq:jacobian_matrix}
    \abs*{\frac{\partial h_j^{(d)}}{\partial x_i} (x_i)} \leq C \cdot (\hat{\bm{A}}^{d})_{ji}
\end{equation}
for a constant $C$. The powers of the normalized augmented adjacency matrix then measure the degree to which a given graph is prone to over-squashing.  In other words, bottlenecks are associated with the entries of the powers of the matrix with small values.  Note that the values cannot be zero, since zero values indicate no edge between two nodes.

It is also important to note that in our setting of graphs the Jacobian~\eqref{eq:jacobian} is computed as a discrete derivative.  In our work, we assume that the Jacobian is computed by numerical approximations; no further details were provided \cite{bottleneck-bronstein}.

There also exist other measures to quantify over-squashing.  For example, the Cheeger constant is a direct measure of over-squashing that captures how easy or difficult it is to totally disconnect a graph; however, it is known to be NP-hard to compute \citep{cheeger}.

\section{Discrete Geometry and Curvature}
\label{sec:math}

In this section, we turn to the mathematical aspects of \emph{discrete curvature}, which may be used to model information flow on a network \citep[e.g.,][]{cgnn,bottleneck-bronstein}.  Here, we present the origins of smooth geometry and curvature, and discuss the evolution towards discrete notions.  Most importantly, we define all discrete curvatures that will be implemented in our study.

%There are several types of discrete curvature. One of the most commonly used ones in research is Ollivier Ricci curvature \cite{ollivier-original}. It captures a lot of structural information, but is highly abstract and algebraic, hence difficult to implement in code and compute efficiently \cite{discrete-curvature}. 1D~Forman \cite{forman-original, forman-curvature}, augmented (2D) Forman \cite{2d-forman} and Haantjes \cite{haantjes2, haantjes} curvatures, however, have concise formulas and are fast to compute. 

\subsection{Ricci Curvature and Ricci Flow}

\begin{figure}
\centering
\begin{subfigure}[t]{0.4\textwidth}
\centering
\includegraphics[width=0.9\linewidth]{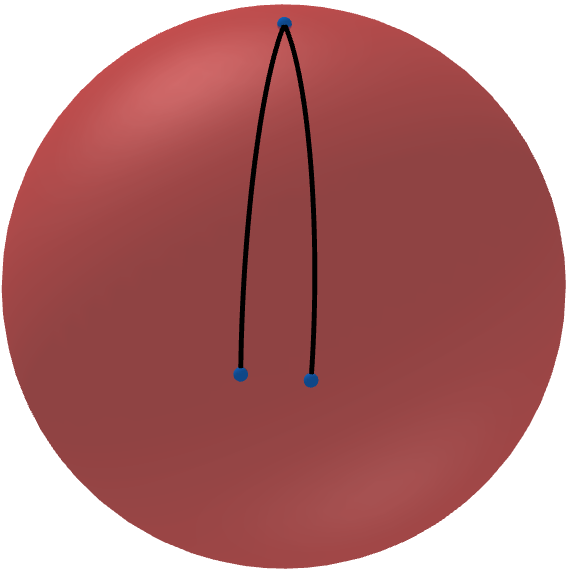}
\caption{$\mathrm{R}(x) > 0$}
\label{fig:sphere1}
\end{subfigure}
\begin{subfigure}[t]{0.4\textwidth}
\centering
\includegraphics[width=0.9\linewidth]{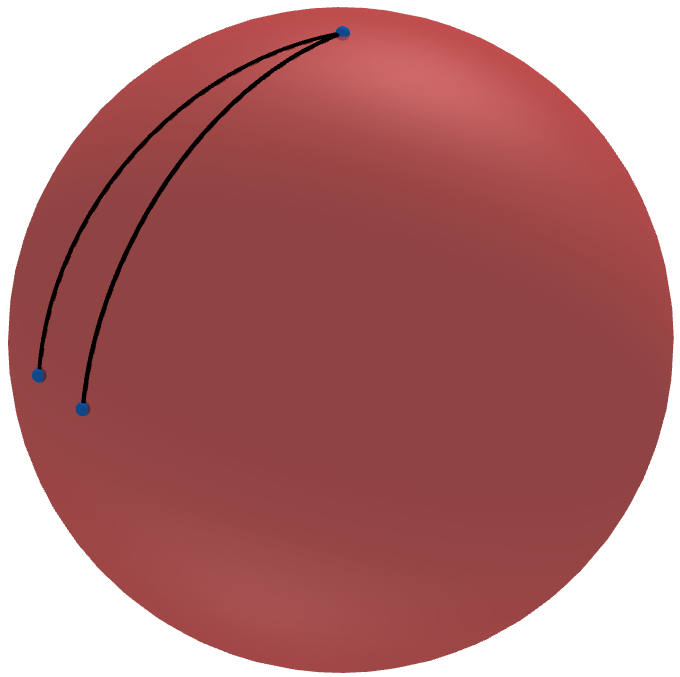}
\caption{$\mathrm{R}(x) > 0$}
\label{fig:sphere2}
\end{subfigure}
\begin{subfigure}[t]{0.4\textwidth}
\centering
\includegraphics[width=0.9\linewidth]{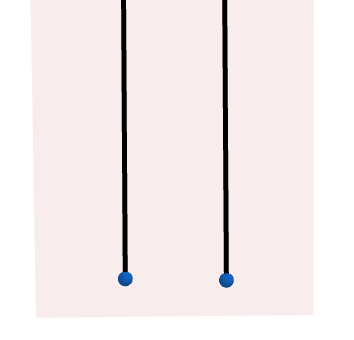}
\caption{$\mathrm{R}(x) = 0$}
\label{fig:plane}
\end{subfigure}
\begin{subfigure}[t]{0.4\textwidth}
\centering
\includegraphics[width=0.9\linewidth]{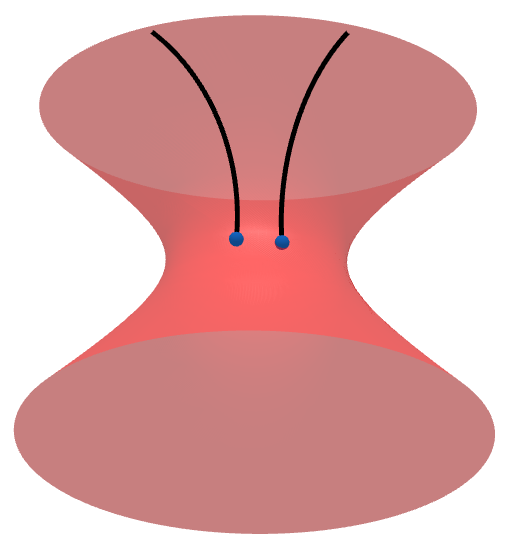}
\caption{$\mathrm{R}(x) < 0$}
\label{fig:hyperbolic}
\end{subfigure}
\caption{View of two geodesics shot in parallel from blue points on example manifolds to illustrate Ricci curvature, $\mathrm{Ric}(x)$. (\subref{fig:sphere1}), (\subref{fig:sphere2}) Two perspectives on a sphere, where the geodesics converge at the top of the sphere. (\subref{fig:plane}) On a plane, the geodesics remain parallel. (\subref{fig:hyperbolic}) On a hyperbolic manifold, the geodesics diverge.}
\label{fig:ricci-examples}
\end{figure}

The \textit{Ricci curvature} of differential geometry is, roughly speaking, a measure that quantifies the extent to which a Riemannian manifold locally differs from a Euclidean space in various tangential directions.  In particular, Ricci curvature determines whether two geodesics shot in parallel from two nearby points on a given manifold tend to converge, remain parallel, or diverge along the manifold.  Then curvature is positive, if the geodesics converge to a single point; zero, if the geodesics remain parallel; and negative, if the geodesics diverge; see Figure~\ref{fig:ricci-examples}.  The quicker the convergence or divergence, the larger the absolute value of Ricci curvature.  

\paragraph{Ricci Flow.}

Ricci curvature can be used to smooth a manifold via the \textit{Ricci flow}, namely the partial differential equation
\begin{equation}
    \frac{\partial \bm{g}}{\partial t} = -2 \rm{Ric}(\bm{g}),
    \label{eq:ricci-flow}
\end{equation}
where $\bm{g}$ denotes the Riemannian metric and $\rm{Ric}$ the Ricci curvature \citep{ricci-flow}.
It should be noted that in most discretizations the 2-dimensional version of the flow is adopted (see e.g. \cite{GuYau}. In this dimension, $\rm{Ric} = K\bm{g}$, where $K$ denotes the classical Gauss curvature, thus the Ricci flow becomes
\begin{equation*}
    \frac{\partial \bm{g}}{\partial t} = -2 K(\bm{g}).
    \label{eq:ricci-flow}
\end{equation*}
In the discrete setting of meshes or networks, the PDE above becomes an ODE, thus the flow is reversible, which a fact of practical importance in many applications and, in particular, the one we study in this paper. Also observe that regions where $K > 0$ ($\rm{Ric} > 0$) tend to shrink, while those with $K < 0$ ($\rm{Ric} < 0$) tend to expand.
%More precisely, what \eqref{eq:ricci-flow} expresses is that at each time step, the metric tensor increases (manifold is expanded) where the Ricci curvature is negative and decreases (manifold is shrunk) where the Ricci curvature is positive, effectively making the manifold "rounder". 

\begin{figure}
\centering
\begin{subfigure}[t]{.49\textwidth}
\centering
\includegraphics[width=0.9\linewidth]{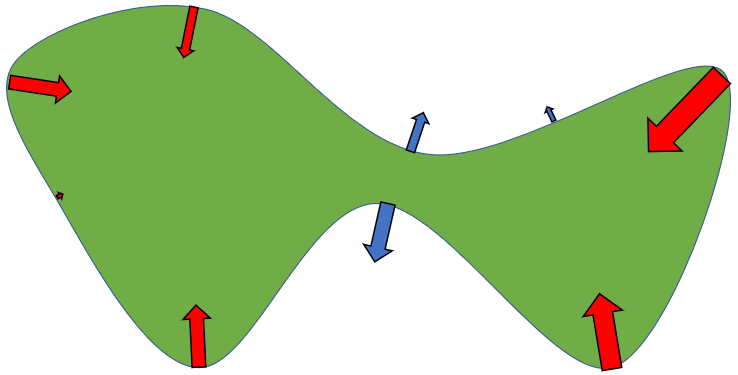}
\caption{}
\label{fig:ricci-flow1}
\end{subfigure}
\begin{subfigure}[t]{.49\textwidth}
\centering
\includegraphics[width=0.9\linewidth]{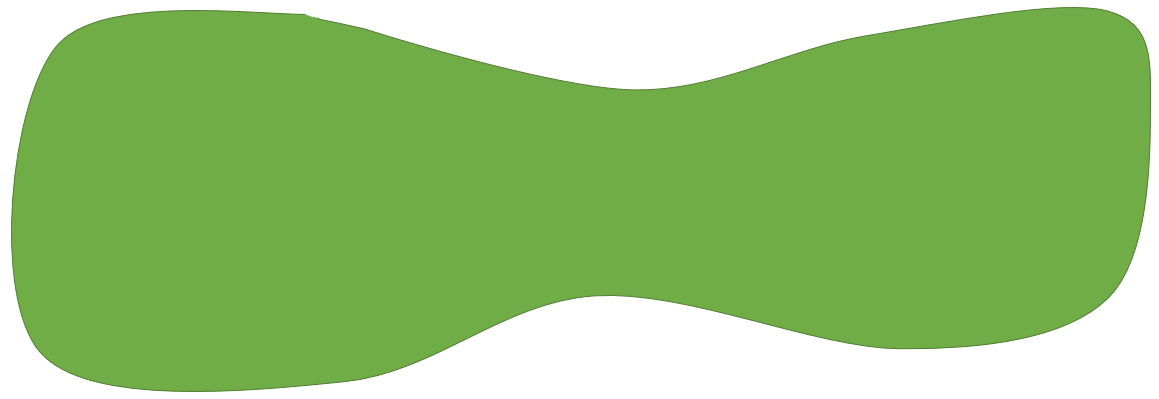}
\caption{}
\label{fig:ricci-flow2}
\end{subfigure}
\caption{Visualisation of the Ricci flow for a 2D shape. (\subref{fig:ricci-flow1}) The arrows show the direction and relative magnitude of the flow at current time step $t$. Red and blue arrows correspond to points with positive and negative curvature respectively, and figuratively represent the metric tensor. (\subref{fig:ricci-flow2}) The manifold at a later time step $t + dt$, expanded and shrunk accordingly to the arrows in (\subref{fig:ricci-flow1}).}
\label{fig:ricci-flow}
\end{figure}

\begin{example}
\label{ex:ricci_flow}
Consider the example manifold in Figure~\ref{fig:ricci-flow} for an intuition on how Ricci flow may be used to smooth a manifold.  In Figure~\ref{fig:ricci-flow1}, the Ricci flow is illustrated by the color and thickness of the arrows and indicate how much as well as the direction in which an expansion produces a smoothed version of the manifold illustrated in Figure~\ref{fig:ricci-flow2}.
\end{example}

In the above Example \ref{ex:ricci_flow}, the regions of negative Ricci curvature where the Ricci flow is illustrated with blue arrows can be seen as a bottleneck of the manifold.  This observation motivates a discretization of the manifolds to graphs, as well as corresponding notions of Ricci curvature and Ricci flow, to be able to use them to model and reduce information over-squashing.

\paragraph{From Manifolds to Graphs.}

In certain instances, there is a natural reduction of manifolds to graphs.  For example, images can be represented in a discrete manner by meshes, which can be seen as 4-regular graphs, while in graphics, data is encoded as triangular meshes whose 1-skeleta are also graphs.  

\begin{figure}
\centering
\begin{subfigure}[t]{0.32\textwidth}
\centering
\includegraphics[width=0.9\linewidth]{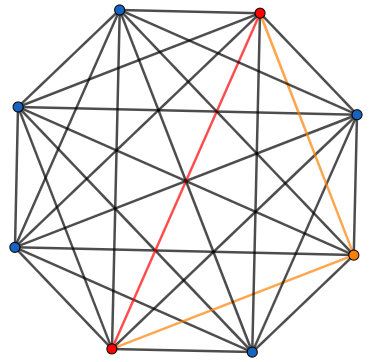}
\caption{}
\label{fig:clique-analogy}
\end{subfigure}
\begin{subfigure}[t]{0.32\textwidth}
\centering
\includegraphics[width=0.9\linewidth]{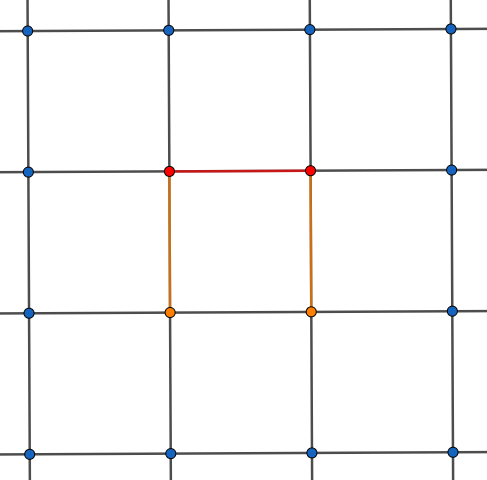}
\caption{}
\label{fig:grid-analogy}
\end{subfigure}
\begin{subfigure}[t]{0.32\textwidth}
\centering
\includegraphics[width=0.9\linewidth]{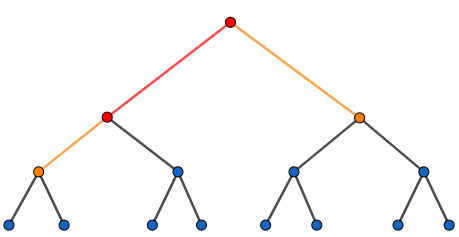}
\caption{}
\label{fig:tree-analogy}
\end{subfigure}
\caption{Graph analogies of manifolds. (\subref{fig:clique-analogy}) A graphical representation of a sphere; (\subref{fig:grid-analogy}) A graphical representation of a plane; (\subref{fig:tree-analogy}) A graphical representation of a hyperbolic manifold.  The example initial pairs nodes and the edges joining them are highlighted in red; the example choices of the next edges and points are highlighted in orange.}
\label{fig:graph-analogies}
\end{figure}

Concretely, for the three types of curvature discussed above (positive, zero, and negative), there exist natural graph analogies.  For a sphere where curvature is positive, a clique is a suitable representation: two parallel geodesics shot from two nearby points on a sphere meet at the top of a sphere, and, likewise, two edges from two adjacent points (connected directly by an edge) in a clique can meet at a common node to create a triangle.
%Of course, edges that do not share an endpoint could be chosen, but the point is to try to choose the "closest" pair of edges, just as only parallel geodesics are considered on a manifold, not any two geodesics. 
For a plane where curvature is flat, a rectangular grid is an appropriate graphical representation: parallel lines on a plane remain parallel forever, and edges from two adjacent points remain parallel. %(at best - it is not possible to find a closer pair of edges; the shortest distance between them is $1$, just like the shortest distance between the initially chosen adjacent nodes).
Finally, a hyperbolic manifold with negative curvature may be represented by a binary tree.  See Figure~\ref{fig:graph-analogies} for graphical examples of manifolds with positive, zero, and negative curvature.
%the parallel geodesics diverge, the edges from two adjacent points also always diverge and the shortest path between them is $3$ (one could say that the distance increased from $1$ to $3$ when trying to move along the tree from the initial two points). This is shown in Figure~\ref{fig:graph-analogies}, and can be compared with respective subfigures of Figure~\ref{fig:ricci-examples}.

\subsection{Discrete Curvature}

With the above intuition of discretizing manifolds to graphs, it is natural to correspondingly define discretized versions of curvature. %Such a discretization is called 
On graphs, discrete curvatures are traditionally node-based measures, however, \textit{discrete Ricci curvature} is an edge-based measure. This is not only natural, given that classical curvature is a directional measure, hence attached to vectors, it also allows for a better and deeper understanding of networks, which are defined by the relationships between their nodes, i.e., by their edges \citep{najman-romon,discrete-curvature}. 

In the discrete Ricci curvature, the edge endpoints correspond to the two nearby points on the manifold from which parallel geodesics are shot to determine the Ricci curvature. We note that the discrete Ricci curvature can also be defined for graph nodes by aggregating, e.g., averaging, the discrete curvature of incident edges, however the notion of node curvature does not play a role in our study of over-squashing which is an edge-specific phenomenon.

There is no single, established definition of discrete curvature. Depending on heuristics, there are many types. Here, we outline the first and best-known discrete curvatures historically proposed for networks. The driving motivation is that, in analogy to the Ricci flow for manifolds, the bottlenecks will have the lowest discrete curvature in the graph.

Note that in our setting, we work with undirected networks and these curvatures are defined for undirected networks.  Analogues for directed networks exist, but since the interest of this work is to explore the role of various discrete curvatures on networks and their performance in reducing over-squashing as studied by \cite{bottleneck-bronstein}, who study the undirected case, we follow suit in our work.  Furthermore, we work with unweighted networks, which give rise to combinatorial properties of graphs that lend computational benefits.

\paragraph{1D Forman curvature.}

\begin{definition}
\label{def:1dforman}
For two nodes $v_1, v_2$ in a graph and an edge $e$ between them, the general \textit{1D Forman curvature} of $e$ is given by \citep{forman-curvature}:

\begin{equation}
    F_{\mathrm{full}}(e) = w_e\left(\frac{w_{v_1}}{w_e} + \frac{w_{v_2}}{w_e} - \sum_{e  _{v_1} \sim e, e_{v_2} \sim e}\left[\frac{w_{v_1}}{\sqrt{w_ew_{e_{v_1}}}} + \frac{w_{v_2}}{\sqrt{w_ew_{e_{v_2}}}}\right] \right),
    \label{eq:forman-1d}
\end{equation}
where $e_{v_1}\sim e$ and $e_{v_2}\sim e$ denote the edges other than $e$ that that are adjacent to nodes $v_1$ and $v_2$ respectively; $w_e$, $w_{e_{v_1}}$, and $w_{e_{v_2}}$ denote the weights of edges $e$, $e_{v_1}$ and $e_{v_2}$ respectively; and $w_{v_1}$ and $w_{v_2}$ denote the weights of the nodes $v_1$ and $v_2$ respectively.
\end{definition}

Recall, however, that here, we study unweighted graphs, which means that only \textit{combinatorial weights} of nodes and edges are considered, i.e., the weights of all nodes and edges are equal to $1$. In this case, \eqref{eq:forman-1d} becomes simply
\begin{equation}
    F(e) = 4 - (\operatorname{deg}(v_1) + \operatorname{deg}(v_2)),
    \label{eq:forman-1d-comb}
\end{equation}
where $\operatorname{deg}(x)$ denotes the degree of node $x$. Note that the first term is $4$ rather than $2$ because the node $v_2$ is counted as a neighbor in $\operatorname{deg}(v_1)$ and vice versa.

In our setting, the 1D Forman curvature is given by \eqref{eq:forman-1d-comb} which is a very simple expression and extremely fast to compute, and is concerned only by the degrees of the endpoints of the edge under consideration. The highest value of the curvature is equal to $2$ and is attained when the edge is disconnected from the rest of the graph. The 1D Forman curvature is negative for the majority of edges in general, as it is always negative when the edge is directly connected to at least $3$ other edges.

The drawback of the simplicity of this curvature is that it is not always very descriptive, even in comparison with the curvature values of other edges in the graph, since in the model case of combinatorial weights, the 1D Forman curvature gives information only about the number of edges directly connected to the edge under consideration. For example, for two clique-like subgraphs connected by one edge, as in Figure~\ref{fig:rewiring1}, the bottleneck would be correctly identified as an edge with the lowest curvature. However, it would generally assign lower curvature to clique-like components of the graph rather than the tree-like components, as an edge in a clique is connected directly to all of the other edges in the clique, while e.g., in a binary tree it is only directly connected to $3$ other edges. Thus the measure of interest is the relative curvature in comparison to the curvature of other edges in the graph, mainly in the combinatorial case.

\paragraph{Augmented Forman curvature.} 
The \textit{augmented Forman curvature} or \textit{2D Forman curvature} attempts to solve the above-mentioned drawbacks of 1D Forman curvature. 

\begin{definition}
For two nodes $v_1, v_2$ in a graph and an edge $e$ between them, the \emph{augmented Forman curvature} or \emph{2D Forman curvature} is given by \citep{2d-forman}:

\begin{equation}
    % F^\#(e) = w_e\left[\left(\sum_{e<f}\frac{w_e}{w_f}+\sum_{v<e}\frac{w_v}{w_e}\right)-\sum_{\hat{e}\parallel e}\left(\sum_{\hat{e},e<f}\frac{\sqrt{w_e \cdot w_\hat{e}}}{w_f} - \sum_{v<e,v<\hat{e}}\frac{w_v}{\sqrt{w_e \cdot w_\hat{e}}}\right)\right]
    F^\#_{\mathrm{full}}(e) = w_e\left[\left(\sum_{e<f}\frac{w_e}{w_f}+\sum_{v<e}\frac{w_v}{w_e}\right)-\sum_{\hat{e}\parallel e}\left\lvert\sum_{\hat{e},e<f} \frac{\sqrt{w_e \cdot w_{\hat{e}}}}{w_f}\sum_{v<e,v<\hat{e}} \frac{w_v}{\sqrt{w_e \cdot w_{\hat{e}}}}\right\rvert\right],
    \label{eq:forman-2d}
\end{equation}
where $a\parallel b$ denotes that $a$ is parallel to $b$, i.e., $a$ and $b$ have a common higher or lower dimensional graph face (e.g., they are two edges that are both a part of the same triangle, which would be in turn equivalent to them having a common neighbour); $a < b$ denotes that $a$ is a graph face of $b$ (e.g., $a$ is an edge and $b$ is a triangle that $a$ is a part of); and the rest of the notation is as in Definition \ref{def:1dforman} (here the faces are also weighted). 
\end{definition}

The mathematical definition of augmented Forman curvature captured in \eqref{eq:forman-2d} is significantly more complex than 1D Forman curvature. However, following \cite{2d-forman}, we can consider solely $3$-cycles, i.e., triangles, and chose again only combinatorial weights. This reduces \eqref{eq:forman-2d} to the following form that relates to \eqref{eq:forman-1d-comb} \citep{2d-forman}
\begin{equation}
    F^\#(e) = F(e) + 3t,
    \label{eq:forman-2d-comb}
\end{equation}
where $t=|N(v_1)\cap N(v_2)|$ is the number of triangles containing the edge $e=(v_1, v_2)$ under consideration.

The idea is that the curvature $F^\#$ \eqref{eq:forman-2d-comb} increases in relation to $F$ if an edge is contained in some triangles. More precisely, the factor of $t$ in \eqref{eq:forman-2d-comb} equal to $3$ guarantees that edges that create a triangle together with the edge under consideration $e$ do not contribute negatively to the curvature, but positively instead. Indeed, there should intuitively be no problem with information over-squashing within a $3$-cycle, which is the simplest form of a clique. 

For each pair of edges that create a triangle with $e$, the curvature increases by $1$, while for the 1D Forman curvature if would simply decrease the curvature by $2$, due to the contribution to the degrees of endpoints of $e$. If an edge is not a member a triangle with $e$, it contributes negatively to the augmented Forman curvature by decreasing it by $1$, just as in the 1D version. Hence, the augmented version maintains a balance between the growth of degrees of endpoints and creation of $3$-cycles.

\paragraph{Haantjes curvature.}
The \textit{Haantjes curvature} \citep{haantjes2} is less common than the Forman curvatures, even though its definition is by far the simplest and most intuitive of all the discrete network curvatures; see \cite{haantjes} for its even simpler network adaptations.

\begin{definition}
\label{def:haantjes}
Consider a graph where all weights are equivalently equal to 1 (i.e., the combinatorial case).  For two nodes $v_1, v_2$ in a graph and an edge $e$ between them, the \emph{Haantjes curvature} is given by 
\begin{equation}
    \kappa_H^2(e) = t,
    \label{eq:haantjes}
\end{equation}
where $t$ is as in \eqref{eq:forman-2d-comb}, i.e., the number of triangles containing the edge $e$.
\end{definition}

The original Haantjes curvature is a metric curvature, thus in the network case it takes into account solely edge weights. However, in practice, one can devise new edge weights that incorporate both the original ones as well as the given vertex weights (see \cite{haantjes}).  Definition \ref{def:haantjes} is commonly used in graphics settings and simply counts the triangles adjacent to a given edge, i.e., the number of $3$-cycles containing the edge under consideration, $e$.

%In the special case at hand, where combinatorial (i.e. $\equiv 1$) weights are considered, Haantjes curvature reduces to an approach to discrete curvatures common in Graphics, for instance, that is to the simple count of triangles adjacent to the given edge, i.e. of $3$-cycles containing the edge under consideration:
%
%\begin{equation}
%    \kappa_H^2(e) = |N(v_1)\cap N(v_2)|,
%    \label{eq:haantjes}
%\end{equation}
%
%where $e=(v_1, v_2)$. 

As a consequence, the Haantjes curvature is indeed higher for clique-like components of a graph than for tree-like components (each edge of a $k$-clique is a part of $k-2$ triangles, but there are no triangles in a tree). Haantjes curvature is trivially nonnegative, which is also in contrast with 1D Forman, where the majority of edges usually have negative curvature. The augmented Forman curvature can now be thought of as a balance between 1D Forman and Haantjes curvatures. (Note that an ``augmented'' Haantjes curvature, that takes into account face weights as well, has been introduced in \cite{haantjes}.)

\paragraph{Balanced Forman curvature.}
The most recent proposal for discrete curvature that will also be studied in this paper is is the \textit{balanced Forman curvature (BFC)}.  It was introduced specifically in the context of over-squashing on graphs \citep{bottleneck-bronstein}. 

\begin{figure}
\centering
\begin{subfigure}[t]{0.32\textwidth}
\centering
\includegraphics[width=0.9\linewidth]{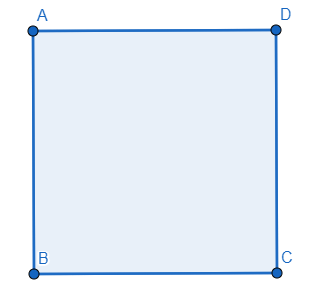}
\caption{$s(B)=s(C)=1$}
\label{fig:4-cycle}
\end{subfigure}
\begin{subfigure}[t]{0.32\textwidth}
\centering
\includegraphics[width=0.9\linewidth]{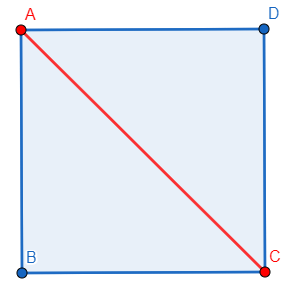}
\caption{$s(B)=s(C)=0$}
\label{fig:4-cycle-diag}
\end{subfigure}
\begin{subfigure}[t]{0.32\textwidth}
\centering
\includegraphics[width=0.9\linewidth]{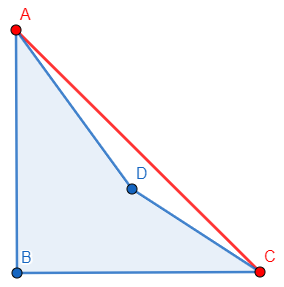}
\caption{$s(B)=s(C)=0$}
\label{fig:4-cycle-diag-degen}
\end{subfigure}
\caption{Values of $s$ for $e=(B, C)$.}
%Note that the $4$-cycles $ABCD$ in (\subref{fig:4-cycle-diag}) and (\subref{fig:4-cycle-diag-degen}) are topologically the same, and both contain the diagonal $AC$.}
\label{fig:4-cycles}
\end{figure}

\begin{definition}[Balanced Forman curvature, \citep{bottleneck-bronstein}]

Consider an edge $e=(v_1, v_2)$.  Let $d_i = \operatorname{deg}(v_i)$ for $i=1,2$; $t=\left\lvert N(v_1)\cap N(v_2)\right\rvert$, the number of triangles containing $e$; $s(v_i)=|\{ k \in N(v_i) \setminus N(v_j), k \neq v_j : \exists \ w \in (N(k) \cap N(v_j)) \setminus N(v_i), w \neq v_i \}|$ for $i,j=1,2$, the number of neighbors of $v_1$ that create a $4$-cycle (square) that contains $e$ and does not contain any diagonals (see Figure~\ref{fig:4-cycles}); and $\gamma_{max}$ be the maximal number of $4$-cycles that contain $e$ traversing a common node. %(to compute it, calculate the number of $4$-cycles containing $e$ and vertex $v$ for each neighbour $v$ of $v_1$ or $v_2$ other than $v_2$ or $v_1$ and take the largest value).
Then the \emph{balanced Forman curvature} is defined as $\mathrm{BFC}(e) := 0$ if $\min(d_1, d_2) = 1$ and 
\begin{equation}
    \mathrm{BFC}(e) := 
    \displaystyle \frac{2}{d_1} + \frac{2}{d_2} - 2 + 2 \frac{t}{\max(d_1, d_2)} + \frac{t}{\min(d_1, d_2)} + \frac{\gamma_{max}^{-1}}{\max(d_1, d_2)} (s(v_1) + s(v_2))
\end{equation}
otherwise.  
\end{definition}

The idea behind the BFC is to preserve a balance between the complexity of computation (in the spirit of the simple formulation of the classical Forman curvature) and the richness of structural information associated with neighboring edges.  In particular, the BFC formulation takes into account $3$- and $4$-cycles, as well as ``loose'' neighboring edges, i.e., those that do not create $3$- or $4$-cycles.  Here, loose edges have a negative curvature contribution, $3$-cycles have a positive curvature contribution, and $4$-cycles have zero curvature contribute to the BFC.  These components are normalized by node degrees.

Note that the BFC is similar to the Forman and Haantjes curvatures where $3$-cycles are explicitly taken into consideration, while $4$-cycles are considered via loose edges.  Here, in contrast to the BFC, loose edges are considered as components of $4$-cycles and have a negative curvature contribution to the Forman and Haantjes curvatures.

%It is clear from the definition that, besides \textit{loose} neighbouring edges (not creating $3$-cycles or $4$-cycles) and $3$-cycles like Forman and Haantjes curvatures, it is additionally concerned with $4$-cycles. Lose edges are supposed to contribute negatively to the curvature, $3$-cycles positively, and $4$-cycles (without diagonals - if there were diagonals, the edges would be interpreted as parts of $3$ cycles instead) not at all (zero contribution, in contrast to Forman and Haantjes curvatures, where the edges being part of $4$-cycles would be interpreted as loose edges and have negative contribution to the curvature). Corresponding parts of the definition are heuristically normalized by node degrees. 

%The BFC, however, is less informative than Ollivier curvature, and still significantly more complex to compute than Forman and Haantjes curvatures. 

\subsection{Discrete Ricci Flow: The Stochastic Discrete Ricci Flow}

We now outline the discretization of Ricci flow that will be implemented in our experimental work, namely, the \emph{stochastic discrete Ricci flow} (SDRF) algorithm \citep{bottleneck-bronstein}.  Specifically, it is a graph rewiring algorithm that was introduced with the aim of addressing the problem of over-squashing in GNN training; it is designed to support edges with low curvature which are identified as the bottlenecks by adding new edges to increase curvature to increase the efficiency of message passing.  It operates very much in the same spirit as Ricci flow, where, in particular, regions of negative or low curvature are identified and compensated by an opposite effect depending on the negativity in order to smooth the manifold.  Additionally, it incorporates a mechanism to prevent a blow-up on the size of the graph.  The algorithm thus intakes a graph and produces another graph where the regions of the most negatively curved edges of the input graph are augmented with additional edges to increase the curvature at those regions.

%The curvature values of edges can be used to determine the lowest-curved ones and rewire the graph using a process called \textit{Stochastic Discrete Ricci Flow} \cite{bottleneck-bronstein} (SDRF), which supports the low-curved edges by adding new edges around them and effectively increasing their curvature, to reduce over-squashing and perform message passing more efficiently.

At each iteration, the algorithm chooses the edge with the smallest curvature, candidate edges to add to support the edge under consideration, and the edge to add from candidates with softmax probability (regulated with a \textit{temperature} parameter $\tau$) with the aim to increase curvature, where this latter value is calculated as the difference between curvature of the edge under consideration before and after adding the support edge. The algorithm then chooses the edge with the highest curvature and, if this curvature value surpasses a certain threshold, removes this edge from the graph and thus ensures a bound on the size of the graph. The process repeats until either the convergence is reached, in the sense that there are no additional candidates and no edges to remove, or the maximum number of iterations is reached. 

\begin{algorithm}[H]
\small
    \caption{Stochastic Discrete Ricci Flow (SDRF)}\label{alg:sdrf}
    \begin{algorithmic}
        \Input: 
            graph $G$, temperature $\tau > 0$, max number of iterations, discrete curvature Curv, optional Curv upper-bound $C^+$
        \EndInput
        \Repeat
            \For{edge $i\sim j$ with minimal discrete curvature $\mathrm{Curv}(i,j)$}
            \State Calcukate vector $\bm{x}$ where $x_{kl}=\mathrm{Curv}_{kl}(i,j)-\mathrm{Curv}(i,j)$, the improvement to $\mathrm{Curv}(i,j)$ from adding edge $k \sim l$ where $k\in N(i) \cup \{i\}, l\in N(j) \cup \{j\}$;
            \State Sample index $k,l$ with probability softmax$(\tau \bm{x})_{kl}$ and add edge $k\sim l$ to $G$.
            \EndFor
            \State Remove edge $i\sim j$ with maximal discrete curvature Curv$(i,j)$ if Curv$(i,j) > C^+$.
            % \Until conv
            % \State Set $e=(v_1, v_2)$ to be the edge with minimal curvature $\mathrm{C}(e)$
            % \State Calculate matrix $\bm{x}$ where $\bm{x}_{kl} = \mathrm{C}_f(e) - \mathrm{C}(e)$, the improvement to $\mathrm{C}(e)$ after adding edge $f=(k, l)$ where $k \in N(v_1) \cup \{ v_1 \}$ and $l\in N(v_2) \cup \{ v_2 \}$
            % \State Sample edge $f$ with probability softmax$(\tau \bm{x})_{kl}$ and add $f$ to $G$.
            % \State Remove the edge $g$ with maximal curvature $\mathrm{C}(g)$ if $\mathrm{C}(g) > C^+$
        \Until convergence or max iterations reached
    \end{algorithmic}
\end{algorithm}

\begin{example}
See Figure~\ref{fig:rewiring} for an example run of Algorithm~\ref{alg:sdrf}.
\end{example}

%Note that when implementing Algorithm~\ref{alg:sdrf} with the BFC in particular, the candidate edges $f$ that are sampled to support the edge with lowest curvature are chosen to be the edges creating $3$-cycles or $4$-cycles.

Notice here that the curvature computation is incorporated into the SDRF algorithm when the softmax probability is computed for each candidate selection.

%%%%%%%%%%%%%%%%%%%%%%%%%%%%%%%%%%%%%%%%%%%%%%%%%%%%%%%%%%%%%%%%%%%%

\section{Data and Experimental Setup}
\label{sec:methods}

In this section, we describe the datasets used and give details on the setup of our experimental study.  The aim is to test the performance in terms of accuracy and computational runtime of various discrete curvatures in the SDRF algorithm that is designed to reduce information over-squashing in training GNNs.  With this aim in mind, the experiments were set up to closely align with the setup of \cite{bottleneck-bronstein} in order to better relate the findings.  Furthermore, to ensure fairness of method evaluation, the performance on additional datasets that were not studied by \cite{bottleneck-bronstein} was also evaluated, resulting in a wider variety of dataset applications and an independent implementation of their proposed BFC.  Recall, however, that an important difference between our work and theirs is that only one curvature---the BFC---was used in their curvature-based rewiring.

\subsection{Datasets}

We used the following 12 benchmarking datasets in our experimental study:
\begin{itemize}
    \item Cora \citep{cora} and Citeseer \citep{Citeseer}\\
    Large citations datasets containing information about presence of specific words in publications
    \item Pubmed \citep{pubmed}\\
    Large citations dataset containing information about diabetes of patients classified into one of three classes
    \item Cornell, Texas, and Wisconsin \citep{ctw}\\
    Small datasets containing information about world wide web pages collected from computer science departments of corresponding universities
    \item Chameleon, Squirrel \citep{chameleon-squirrel}, and Actor \citep{actor}\\
    Large datasets based on the Wikipedia networks
    \item Computers and Photo \citep{computers-photo}\\
    Large e-commerce (Amazon) datasets,
    \item Coauthor CS \citep{cgnn}\\
    Large citation dataset with papers in computer science
\end{itemize}
The datasets Computers, Photo and Coauthor CS were not evaluated in \cite{bottleneck-bronstein}. The details of the datasets are summarized in Table~\ref{table:datasets}.

\begin{table}[h]
\centering
 \begin{tabular}{|c|c c c c c c|} 
 \hline
 & Cora & Citeseer & Pubmed & Cornell & Texas & Wisconsin\\ \hline%[0.5ex] 
%  \hline\hline
 Nodes & 2485 & 2120 & 19717 & 140 & 135 & 184\\
 Edges & 5069 & 7358 & 44324 & 219 & 251 & 362\\
 Features & 1433 & 778 & 500 & 1703 & 1703 & 1703\\
 Classes & 7 & 6 & 3 & 5 & 5 & 5\\
 Undirected & Yes & Yes & Yes & No & No & No\\% [1ex] 
 \hline
 \end{tabular}
 
 \br
 
 \begin{adjustbox}{max width=\textwidth}
 \begin{tabular}{|c | c c c c c c|} 
 \hline
 & Chameleon & Squirrel & Actor & Computers & Photo & Coauthor CS\\ %[0.5ex] 
 \hline
 Nodes & 832 & 2186 & 4388 & 13381 & 7487 & 18333\\ 
 Edges & 12355 & 65224 & 21907 & 245778 & 119043 & 81894\\
 Features & 2323 & 2089 & 931 & 767 & 745 & 6805\\
 Classes & 5 & 5 & 5 & 10 & 8 & 15\\
 Undirected & No & No & No & Yes & Yes & Yes\\ %[1ex] 
 \hline
 
 \end{tabular}
 \end{adjustbox}
 \caption{Details of datasets.  `Undirected' specifies whether the network is undirected by default.  If not, it is made undirected for the experiments.}
 \label{table:datasets}
\end{table}

\subsection{Experimental Design}

We tested the performances of no curvature (i.e., no rewiring); 1D Forman curvature; augmented Forman curvature; Haantjes curvature; and balanced Forman curvature in the SDRF algorithm for graph rewiring.  The implementation of the SDRF algorithm was taken from the repository associated with \cite{bottleneck-bronstein} available at \url{https://github.com/jctops/understanding-oversquashing}.  Other design choices and setup parameters such as such as data loading, selection of largest connected component, network type, hyperparameters, and seeds have been set following \cite{bottleneck-bronstein}; Table~\ref{table:hyperparams} presents the hyperparameters used for training of GNN models. 

\begin{table}[h]
\centering
\begin{adjustbox}{max width=0.95\textwidth}
 \begin{tabular}{|c| c c c c c c|} 
 \hline
 & Cora & Citeseer & Pubmed & Cornell & Texas & Wisconsin\\ %[0.5ex] 
 \hline
 Dropout & 0.3396 & 0.4103 & 0.3749 & 0.2911 & 0.216 & 0.2452\\ 
 Hidden depth & 1 & 1 & 3 & 1 & 1 & 1\\
 Hidden dimension & 128 & 64 & 128 & 128 & 64 & 64\\
 Learning date & 0.0244 & 0.0199 & 0.0112 & 0.0056 & 0.0229 & 0.2452\\
 Weight decay & 0.1076 & 0.4551 & 0.0138 & 0.0366 & 0.0137 & 0.1559\\
 Max iterations & 100 & 84 & 166 & 126 & 89 & 136\\
 $\tau$ & 163 & 180 & 115 & 145 & 22 & 12\\
 Removal bound & 0.95 & 0.22 & 14.43 & 0.88 & 1.64 & 7.95\\
 Patience & 100 & 10 & 10 & 100 & 100 & 100\\ %[1ex] 
 \hline
 \end{tabular}
 \end{adjustbox}
 
 \br
 
 \begin{adjustbox}{max width=0.95\textwidth}
 \begin{tabular}{|c | c c c c c c|} 
 \hline
 & Chameleon & Squirrel & Actor & Computers & Photo & Coauthor CS\\ %[0.5ex] 
 \hline
 Dropout & 0.4886 & 0.3079 & 0.3424 & 0.3396 & 0.3396 & 0.3396\\ 
 Hidden depth & 1 & 1 & 1 & 1 & 1 & 1\\
 Hidden dimension & 32 & 32 & 64 & 128 & 128 & 128\\
 Learning rate & 0.0268 & 0.0299 & 0.0129 & 0.0244 & 0.0244 & 0.0244\\
 Weight decay & 0.4056 & 0.0158 & 0.0126 & 0.1076 & 0.1076 & 0.1076\\
 Max iterations & 2442 & 1396 & 3249 & 100 & 100 & 100\\
 $\tau$ & 252 & 436 & 106 & 163 & 163 & 163\\
 Removal bound & 2.84 & 5.88 & 0 & 0.95 & 0.95 & 0.95\\
 Patience & 10 & 10 & 10 & 10 & 10 & 10\\ %[1ex] 
 \hline
 
 \end{tabular}
 \end{adjustbox}
 \caption{Training hyperparameters.  Note here that $\tau=163$ for the Squirrel dataset for rewiring using augmented Forman curvature (same $\tau$ as for Cora), since for $\tau=436$ we encounter integer overflow.}
 \label{table:hyperparams}
\end{table}

\paragraph{Software and Data Availability.}
The full implementation of the SDRF algorithm with all curvatures studied incorporated and datasets are freely and publicly available at \url{https://github.com/jakubbober/discrete-curvature-rewiring}.

%%%%%%%%%%%%%%%%%%%%%%%%%%%%%%%%%%%%%%%%%%%%%%%%%%%%%%%%%%%%%%%%%%%%
\section{Results: Supervised Learning with Graph Rewiring}
\label{sec:results}

We now present the results of the supervised learning task of SDRF-based graph rewiring on each of the 12 datasets discussed in the previous section.  We report results on accuracy and computational runtime.

\subsection{Accuracy}

Each experiment was run for 100 seeds, so we report 95\% confidence intervals of mean accuracies are reported using a $z$-score of 1.96.  For reference and performance comparison, the 95\% confidence intervals for the SDRF-rewiring using BFC reported by  \cite{bottleneck-bronstein} are also given for those relevant datasets. 

%Multiple variations of experiments are performed. Initially, the \texttt{redo\_rewiring} flag is set to \texttt{False} for both training and testing, as performing separate rewiring for each seed for all datasets would take an unreasonable amount of time. The next experiment has the \texttt{redo\_rewiring} flag set to \texttt{True} and computes results only for datasets for which the computation time is reasonable.

%For reference, the 95\% confidence intervals for the SDRF rewiring using BFC after making the graph undirected reported in \cite{bottleneck-bronstein} are given for the datasets used in \cite{bottleneck-bronstein}. For this experiment, there is only one rewiring instance for all seeds for each dataset and discrete curvature type. The results in \cite{bottleneck-bronstein} tend to be worse for all datasets when not making the graphs undirected regardless, so making the graphs undirected should not really be considered a limitation.

The best two results are highlighted for each dataset in each accuracy table: the best one in red bold, the second best in black bold (excluding the reported BFC results from \cite{bottleneck-bronstein} for reference). The \textit{None} curvature row represents results without any rewiring. OOM indicates that the out of memory error has occurred. N/A in the reference BFC row for Computers, Photo and Coauthor CS datasets indicates that there are no reference results for these datasets as these datasets were not studied by \cite{bottleneck-bronstein}.

\begin{table}
    \centering
    \begin{adjustbox}{max width=\textwidth}
    \begin{tabular}{|l| l l l l l l|}
    \hline
        ~ & Cora & Citeseer & Pubmed & Cornell & Texas & Wisconsin \\ \hline
        None & \textbf{\textcolor{red}{81.65 $\pm$ 0.25}} & 72.14 $\pm$ 0.31 & 77.74 $\pm$ 0.40 & 48.50 $\pm$ 0.60 & 59.19 $\pm$ 0.38 & 50.24 $\pm$ 0.54 \\ 
        1D & 81.15 $\pm$ 0.24 & \textbf{72.17 $\pm$ 0.28} & \textbf{\textcolor{red}{77.76 $\pm$ 0.37}} & 52.75 $\pm$ 0.82 & \textbf{64.59 $\pm$ 1.11} & \textbf{52.70 $\pm$ 0.71} \\ 
        Augmented & \textbf{81.56 $\pm$ 0.24} & 72.12 $\pm$ 0.30 & 77.70 $\pm$ 0.40 & 55.43 $\pm$ 0.62 & \textbf{\textcolor{red}{65.48 $\pm$ 1.23}} & 52.62 $\pm$ 0.74 \\ 
        Haantjes & 81.55 $\pm$ 0.25 & \textbf{\textcolor{red}{72.19 $\pm$ 0.30}} & \textbf{77.75 $\pm$ 0.38} & \textbf{56.29 $\pm$ 0.50} & 63.33 $\pm$ 0.94 & \textbf{\textcolor{red}{55.81 $\pm$ 0.77}} \\ 
        BFC & 81.38 $\pm$ 0.25 & 72.09 $\pm$ 0.28 & OOM & \textbf{\textcolor{red}{58.39 $\pm$ 0.64}} & 61.11 $\pm$ 0.68 & 48.86 $\pm$ 0.91 \\ \hline
        Reference BFC & 82.76 $\pm$ 0.23 & 72.58 $\pm$ 0.20 & 79.10 $\pm$ 0.11 & 57.54 $\pm$ 0.34 & 70.35 $\pm$ 0.60 & 61.55 $\pm$ 0.86 \\ \hline
    \end{tabular}
    \end{adjustbox}
    
    \br

    \centering
    \begin{adjustbox}{max width=\textwidth}
    \begin{tabular}{|l|l l l l l l|}
    \hline
        ~ & Chameleon & Squirrel & Actor & Computers & Photo & Coauthor CS \\ \hline
        None & \textbf{\textcolor{red}{47.38 $\pm$ 0.45}} & \textbf{\textcolor{red}{38.16 $\pm$ 0.32}} & 27.82 $\pm$ 0.24 & 41.74 $\pm$ 1.41 & \textbf{56.4 $\pm$ 2.85} & \textbf{\textcolor{red}{90.89 $\pm$ 0.11}} \\ 
        1D & 44.88 $\pm$ 0.43 & 36.83 $\pm$ 0.27 & \textbf{29.41 $\pm$ 0.26} & 42.24 $\pm$ 1.58 & 54.93 $\pm$ 3.46 & 90.83 $\pm$ 0.11 \\ 
        Augmented & 43.54 $\pm$ 0.88 & 36.75 $\pm$ 0.25 & \textbf{\textcolor{red}{29.81 $\pm$ 0.30}} & \textbf{42.93 $\pm$ 1.56} & 54.44 $\pm$ 3.01 & \textbf{\textcolor{red}{90.89 $\pm$ 0.11}} \\ 
        Haantjes & 46.14 $\pm$ 0.55 & 36.59 $\pm$ 0.26 & 29.36 $\pm$ 0.24 & \textbf{\textcolor{red}{42.95 $\pm$ 1.74}} & \textbf{\textcolor{red}{56.74 $\pm$ 3.12}} & 90.86 $\pm$ 0.10 \\ 
        BFC & \textbf{46.92 $\pm$ 0.73} & \textbf{37.82 $\pm$ 0.36} & 29.11 $\pm$ 0.25 & 41.55 $\pm$ 1.91 & 54.29 $\pm$ 3.13 & OOM \\ \hline
        Reference BFC & 44.46 $\pm$ 0.17 & 37.67 $\pm$ 0.23 & 28.35 $\pm$ 0.06 & N/A & N/A & N/A \\ \hline
    \end{tabular}
    \end{adjustbox}
    
    \br

    \centering
    \begin{adjustbox}{max width=\textwidth}
    \begin{tabular}{|l|l l l l l l|}
    \hline
        ~ & Cora & Citeseer & Pubmed & Cornell & Texas & Wisconsin  \\ \hline
        None & 81.55 $\pm$ 0.23 & 72.21 $\pm$ 0.29 & \textbf{\textcolor{red}{77.90 $\pm$ 0.36}} & 48.11 $\pm$ 0.60 & 59.33 $\pm$ 0.40 & 49.95 $\pm$ 0.49   \\ 
        1D & 81.10 $\pm$ 0.24 & \textbf{\textcolor{red}{72.45 $\pm$ 0.29}} & \textbf{\textcolor{red}{77.90 $\pm$ 0.35}} & 51.00 $\pm$ 0.88 & \textbf{\textcolor{red}{68.07 $\pm$ 1.09}} & 54.51 $\pm$ 0.84 \\ 
        Augmented & \textbf{\textcolor{red}{81.57 $\pm$ 0.25}} & \textbf{72.22 $\pm$ 0.27} & 77.89 $\pm$ 0.38 & 53.89 $\pm$ 0.63 & \textbf{64.81 $\pm$ 1.20} & \textbf{\textcolor{red}{56.49 $\pm$ 0.79}} \\ 
        Haantjes & \textbf{81.56 $\pm$ 0.24} & 72.10 $\pm$ 0.28 & 77.71 $\pm$ 0.41 & \textbf{\textcolor{red}{57.18 $\pm$ 0.57}} & 64.78 $\pm$ 1.15 & 55.86 $\pm$ 0.76 \\ 
        BFC & 81.25 $\pm$ 0.25 & 72.04 $\pm$ 0.29 & OOM & \textbf{54.61 $\pm$ 0.50} & 58.37 $\pm$ 0.67 & \textbf{56.19 $\pm$ 0.84}\\ \hline
        Reference BFC & 82.76 $\pm$ 0.23 & 72.58 $\pm$ 0.20 & 79.10 $\pm$ 0.11 & 57.54 $\pm$ 0.34 & 70.35 $\pm$ 0.60 & 61.55 $\pm$ 0.86 \\ \hline
    \end{tabular}
    \end{adjustbox}

    \br

    \centering
    \begin{adjustbox}{max width=\textwidth}
    \begin{tabular}{|l|l l l l l l|}
    \hline
        ~ & Chameleon & Squirrel & Actor & Computers & Photo & Coauthor CS \\ \hline
        None & \textbf{\textcolor{red}{46.86 $\pm$ 0.44}} & \textbf{\textcolor{red}{38.25 $\pm$ 0.33}} & 27.69 $\pm$ 0.22 & \textbf{\textcolor{red}{42.45 $\pm$ 1.55}} & 53.39 $\pm$ 2.75 & \textbf{\textcolor{red}{90.90 $\pm$ 0.10}}  \\ 
        1D & 44.99 $\pm$ 0.40 & 36.49 $\pm$ 0.29 & \textbf{29.66 $\pm$ 0.26} & 41.11 $\pm$ 1.86 & \textbf{55.57 $\pm$ 3.14} & 90.82 $\pm$ 0.10  \\ 
        Augmented & 42.69 $\pm$ 0.65 & 36.70 $\pm$ 0.26 & \textbf{\textcolor{red}{29.98 $\pm$ 0.25}} & 41.97 $\pm$ 1.71 & \textbf{\textcolor{red}{56.19 $\pm$ 2.96}} & \textbf{\textcolor{red}{90.90 $\pm$ 0.12}} \\ 
        Haantjes & 45.97 $\pm$ 0.51 & 36.83 $\pm$ 0.24 & 29.52 $\pm$ 0.21 & \textbf{42.38 $\pm$ 1.60} & 55.34 $\pm$ 2.93 & 90.88 $\pm$ 0.11 \\ 
        BFC & \textbf{46.62 $\pm$ 0.70} & \textbf{37.61 $\pm$ 0.34} & 29.34 $\pm$ 0.28 & 42.11 $\pm$ 1.65 & 54.51 $\pm$ 2.89 & OOM \\ \hline
        Reference BFC & 44.46 $\pm$ 0.17 & 37.67 $\pm$ 0.23 & 28.35 $\pm$ 0.06 & N/A & N/A & N/A \\ \hline
    \end{tabular}
    \end{adjustbox}
    
    \caption{95\% confidence intervals of mean accuracies for given datasets and curvature types given in percentages of two experimental runs (first two tables for the first run, last two for the second).}
    \label{table:accuracies}
\end{table}

The results reported in Table~\ref{table:accuracies} indicate that SDRF rewiring generally increases the training performance. The reference BFC results reported in \cite{bottleneck-bronstein} are also generally comparable to the BFC results of the performed experiment, although we note a tendency for our computation of BFC-based SDRF rewiring to be on the lower side, though still in the general range where we are able to claim reproducibility.

In particular, we note that performance for the classical curvatures is generally better than the performance without any rewiring, and often better than performance of BFC. For some results in Table~\ref{table:accuracies}, the simplest form of curvature---the 1D Forman curvature---tends to give the best results. This tends to indicate that the edges with large sums of degrees are the graph bottlenecks and suffer from over-squashing. The results for Haantjes curvature are the best for some of the other datasets, which suggests that association with many $3$-cycles helps an edge to reduce over-squashing. Although with less frequency, the augmented Forman curvature also yields best results for certain experiments, which could mean that maintaining the balance between the two metrics can reduce over-squashing most effectively.

Note, however, that the experiments upon rerunning yielded results that differ quite significantly, especially for small datasets (Cornell, Texas, Wisconsin). For example, Table~\ref{table:accuracies} shows that the Haantjes curvature seems to generally bring the best results in the first run, while the augmented Forman curvature performs best in the second run. More importantly, it is often the case that the corresponding results (dataset--curvature pairs) for different rewirings for the two runs are often not within the respective $95\%$ confidence intervals, indicating a lack of robustness of the results.  One explanation for this phenomenon can be overfitting of the average accuracy to one instance of the SDRF rewiring. This can have a significant impact on the average performance, especially for the small datasets, for which the rewiring of multiple edges can have a greater impact on the graph structure than for larger datasets. The results for these datasets also differ significantly between each type of curvature, and in relation to performance without any rewiring. Moreover, the BFC results differ more significantly for these datasets than for others with respect to the reference BFC results.

%As just stated, the reason may be that each addition or deletion of an edge has a bigger impact on the graph structure, as there are fewer edges. Looking at Table~\ref{table:datasets} in Appendix~\ref{appendix:datasets} and Table~\ref{table:hyperparams} in Appendix~\ref{appendix:hyperparams}, it can be seen that the ratio of maximum iterations to the number of edges is very high.

To further investigate the intuition that adding or deleting edges on smaller graphs impact the overall graph structure more significantly (taking into account that the hyperparameters set in Table~\ref{table:hyperparams} are very high relative to graph size), we re-ran experiments for Cora, Citeseer, Cornell, Texas and Wisconsin datasets with rewiring for each seed. These datasets were selected given that rewiring was the fastest (as will be discussed further on in discussing computational runtime). %The computations for other datasets with rewiring for each seed would take an unreasonable amount of time. 
The test results of these experiments are shown in Table~\ref{table:accuracies-rewiring}. 

\begin{table}
    \centering
    \begin{adjustbox}{max width=\textwidth}
    \begin{tabular}{|l|l l l l l|}
    \hline
        ~ & Cora & Citeseer & Cornell & Texas & Wisconsin \\ \hline
        None & 81.63 $\pm$ 0.24 & 72.13 $\pm$ 0.29 & 48.04 $\pm$ 0.60 & 59.74 $\pm$ 0.36 & 50.11 $\pm$ 0.53  \\ 
        1D & 81.15 $\pm$ 0.26 & \textbf{72.14 $\pm$ 0.31} & 53.39 $\pm$ 0.81 & \textbf{\textcolor{red}{67.00 $\pm$ 1.28}} & \textbf{55.54 $\pm$ 0.89}  \\ 
        Augmented & \textbf{\textcolor{red}{81.64 $\pm$ 0.25}} & 72.05 $\pm$ 0.29 & \textbf{54.93 $\pm$ 0.59} & \textbf{64.56 $\pm$ 1.15} & 55.49 $\pm$ 0.82 \\ 
        Haantjes & \textbf{\textcolor{red}{81.64 $\pm$ 0.24}} & \textbf{\textcolor{red}{72.19 $\pm$ 0.33}} & \textbf{\textcolor{red}{56.50 $\pm$ 0.60}} & 62.96 $\pm$ 0.92 & \textbf{\textcolor{red}{55.95 $\pm$ 0.72}} \\ 
        BFC & 81.18 $\pm$ 0.27 & 72.12 $\pm$ 0.29 & 53.07 $\pm$ 0.74 & 59.19 $\pm$ 0.69 & 54.24 $\pm$ 0.93 \\ \hline
        Reference BFC & 82.76 $\pm$ 0.23 & 72.58 $\pm$ 0.20 & 57.54 $\pm$ 0.34 & 70.35 $\pm$ 0.60 & 61.55 $\pm$ 0.86 \\ \hline

    \end{tabular}
    \end{adjustbox}
    
    \br
    
    \centering
    \begin{adjustbox}{max width=\textwidth}
    \begin{tabular}{|l|l l l l l|}
    \hline
        ~ & Cora & Citeseer & Cornell & Texas & Wisconsin\\ \hline
        None & \textbf{81.56 $\pm$ 0.23} & \textbf{72.24 $\pm$ 0.29} & 48.46 $\pm$ 0.56 & 59.48 $\pm$ 0.40 & 49.97 $\pm$ 0.52 \\ 
        1D & 81.24 $\pm$ 0.23 & \textbf{\textcolor{red}{72.30 $\pm$ 0.29}} & 52.75 $\pm$ 0.80 & \textbf{\textcolor{red}{67.74 $\pm$ 1.26}} & \textbf{55.62 $\pm$ 0.79} \\ 
        Augmented & \textbf{\textcolor{red}{81.69 $\pm$ 0.22}} & 72.23 $\pm$ 0.30 & \textbf{55.39 $\pm$ 0.68} & \textbf{64.93 $\pm$ 1.10} & 55.27 $\pm$ 0.79\\ 
        Haantjes & 81.49 $\pm$ 0.24 & 72.21 $\pm$ 0.27 & \textbf{\textcolor{red}{55.61 $\pm$ 0.58}} & 63.11 $\pm$ 1.05 & \textbf{\textcolor{red}{56.08 $\pm$ 0.82}} \\ 
        BFC & 81.07 $\pm$ 0.25 & 72.01 $\pm$ 0.32 & 53.00 $\pm$ 0.73 & 60.30 $\pm$ 0.80 & 54.59 $\pm$ 0.88 \\ \hline
        Reference BFC & 82.76 $\pm$ 0.23 & 72.58 $\pm$ 0.20 & 57.54 $\pm$ 0.34 & 70.35 $\pm$ 0.60 & 61.55 $\pm$ 0.86 \\ \hline
    \end{tabular}
    \end{adjustbox}
    \caption{95\% confidence intervals of selected datasets with graph rewiring for each seed given in percentages run twice.}
    \label{table:accuracies-rewiring}
\end{table}

Table~\ref{table:accuracies-rewiring} presents results from two runs with rewiring for every seed, which are shown to be significantly more robust. The sizes of the $95\%$ confidence intervals are comparable to those reported previously in Table~\ref{table:accuracies}, but only two pairs of corresponding runs are not contained in the $95\%$ confidence intervals of one another (namely, Cornell--Haantjes and Texas--BFC). As there are $5 \cdot 5=25$ dataset--curvature pairs for which the experiments were run, the mean results are indeed robust and it is reasonable to consider the results as close to being independent and identically distributed (i.i.d.): the probability that two or more out of 25 means of i.i.d.~random variables are not within the corresponding $95\%$ confidence intervals is $1 - 25 \cdot 0.05 \cdot 0.95^{24}\approx 0.635=63.5\%$, which is high.

%In contrast to the results from Table~\ref{table:accuracies-rewiring} with redoing the rewiring, as already stated, the results in Table~\ref{table:accuracies} are not very robust. 
%There are a lot of corresponding pairs (dataset--curvature) of entries within the two runs that are not within the confidence intervals of each other. However, for no rewiring, the differences are minimal and the corresponding pairs usually align within the $95\%$ confidence intervals. This indicates that the runs without rewiring are far from being identically distributed, and hence, as previously suggested, shows that the experiments without redoing the rewiring for every seed generally overfit to the concrete rewiring of a graph for which the model is trained.

Furthermore, the results of these additional experiments are significantly worse than the reference BFC results. This is likely due to the accuracies for differently rewired graphs having been averaged out, as opposed to using the rewiring with the best validation accuracy for the benchmarking. In contrast, the results in Table~\ref{table:accuracies} are slightly better for some dataset--curvature pairs than the reference BFC results, and sometimes slightly worse. When actually using the framework in practice, for the best results, the training can be performed for several different seeds and the model with the best validation accuracy can be chosen with the most effective rewired graph structure.

%From the results in Table~\ref{table:accuracies-rewiring}, slightly different conclusions can be drawn than from the results in Table~\ref{table:accuracies}. For results in Table~\ref{table:accuracies-rewiring}, the best two curvatures for each dataset are almost the same for the two runs, which again shows the robustness of the results. More importantly, the differences in best accuracies for different curvatures are outside the $95\%$ confidence intervals of one another for some datasets, or barely within them (this is especially the case for Texas and Cornell datasets), which means that some curvature types are clearly better for them than others. The exceptions are Cora and Citeseer, where the best two curvatures slightly differ, but they are all confidently within the $95\%$ confidence intervals of one another, and the differences between the accuracies are generally small across the curvatures.

We summarize the test results for rewiring instances and model parameters pairs that achieved the best validation accuracy in the experiments reported in the second run from Table~\ref{table:accuracies-rewiring} in  Table~\ref{table:best-rewirings}.  Only the second run is considered, but this does not have a significantly negative impact on the robustness of the results, since, as previously justified, the results in Table~\ref{table:accuracies-rewiring} are robust. 

\begin{table}
    \centering
    \begin{adjustbox}{max width=\textwidth}
    \begin{tabular}{|l|l l l l l|}
    \hline
        ~ & Cora & Citeseer & Cornell & Texas & Wisconsin \\ \hline
        None & 82.34 & \textbf{\textcolor{red}{74.19}}  & 50.0  & 51.85  & 51.35 \\ 
        1D & 82.23 & 70.48  & \textbf{\textcolor{red}{60.71}}  & \textbf{\textcolor{red}{74.07}}  & 54.05\\ 
        Augmented & \textbf{83.35}  & \textbf{\textcolor{red}{74.19}}  & \textbf{57.14}  & \textbf{\textcolor{red}{74.07}}  & \textbf{\textcolor{red}{59.46}} \\
        Haantjes & \textbf{\textcolor{red}{83.55}}  & 73.23  & 53.57  & 66.67  & \textbf{\textcolor{red}{59.46}}\\ 
        BFC & 82.84  & 74.03  & 56.94  & 62.96  & 54.05  \\ \hline
        Reference BFC & 82.76 & 72.58 & 57.54 & 70.35 & 61.55 \\ \hline
    \end{tabular}
    \end{adjustbox}
    \caption{Accuracy in \% for the best rewiring instances and models from experiments presented in Table~\ref{table:accuracies-rewiring}.}
    \label{table:best-rewirings}
\end{table}

The main conclusion we draw from these experiemnts is that there is no clear curvature type that has better mean performance overall across the multiple datasets, but it is reasonable to conclude that using the classical curvatures for SDRF-based rewiring can lead to significant performance improvement, often achieving better results than BFC. For every dataset, performing the SDRF-based rewiring almost always yields the best test accuracy when using one of the three classical curvature types, compared to BFC (although no rewiring may also yield the best results).  Often, the two best test accuracies are achieved using classical curvatures.

\paragraph{Determining Bottleneckness: Jacobian Bounds.}

We now turn to an assessment of the bottleneckness of each of our datasets as a further validation of our accuracy conclusions.  As described above in Section~\ref{sec:gnn}, heuristically, bottleneckness is more severe the faster the decrease of the minimum nonzero values of the powers of the normalized augmented adjacency matrix \eqref{eq:jacobian_matrix}.

\begin{figure}
\centering
\includegraphics[scale=0.4]{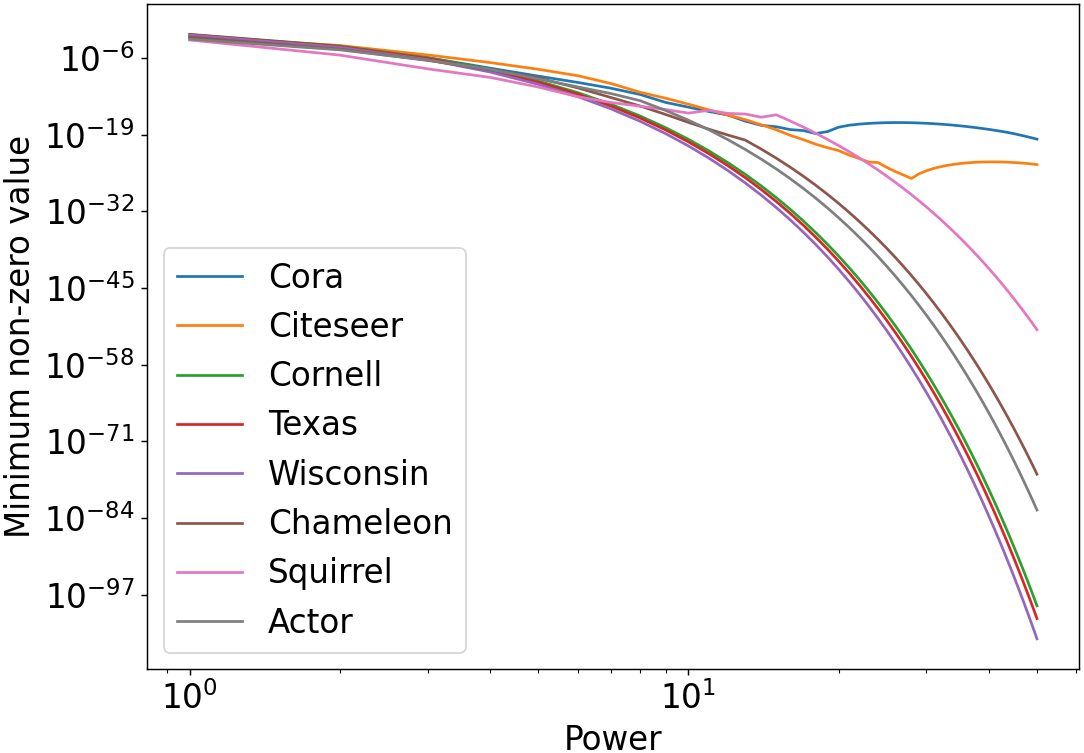}
\caption{Log-log plot of minimum non-zero values of powers of normalized augmented adjacency matrix~\eqref{eq:jacobian_matrix}}
\label{fig:powers}
\end{figure}

Figure~\ref{fig:powers} presents the log-log plot of the minimum non-zero entries of the normalized augmented adjacency matrix for the first $50$ powers for most of the studied datasets. Pubmed, Computers, Photo and Coauthor CS, which are not in this plot, produced adjacency matrices that were too large, which led to killing the processes responsible for computing the powers.

From this figure, we see that the decay of the values is the slowest for Cora and Citeseer datasets. This confirms our previous conclusions, since according to the previously reported results, the difference in accuracy between no rewiring and rewirings for different curvatures was minimal for these datasets suggesting that these datasets are not significantly affected by over-squashing. Similar conclusions can be drawn for the Squirrel dataset, for which the plot in Figure~\ref{fig:powers} also decreases more slowly (especially between $d=10$ and $d=15$), and which is also reported to have the best accuracy for no rewiring in Table~\ref{table:accuracies}.

To capture the reduction of over-squashing via rewiring for a graph, we ran experiments to compute the powers of matrix values as above for both before and after the rewiring. The results of these experiments for several datasets, curvature types, and matrix powers are presented in Table~\ref{table:powers}.

\begin{table}
    \centering
    \begin{adjustbox}{max width=\textwidth}
    \begin{tabular}{|c|c|c c c c c c c c| } 
        \hline
        Curvature & Power & Cora & Citeseer & Cornell & Texas & Wisconsin & Chameleon & Squirrel & Actor \\ \hline
        \multirow{4}{5em}{None}
        & 5 & 8.69e-10 & 1.25e-08 & 1.29e-10 & 7.84e-11 & 3.55e-11 & 4.24e-10 & 1.23e-11 & 2.99e-10 \\
        & 10 & 4.74e-15 & 1.53e-14 & 1.67e-20 & 6.14e-21 & 1.26e-21 & 1.25e-17 & 4.63e-16 & 2.94e-17 \\
        & 20 & 1.86e-18 & 2.09e-22 & 2.79e-40 & 3.77e-41 & 1.59e-42 & 2.35e-31 & 1.49e-21 & 8.62e-34 \\
        & 40 & 8.09e-19 & 2.48e-24 & 7.78e-80 & 1.42e-81 & 2.53e-84 & 5.51e-62 & 2.21e-42 & 7.44e-67 \\ \hline
        \multirow{4}{5em}{1D}
        & 5 & 3.33e-09 & 1.77e-08 & 2.34e-06 & 2.30e-06 &  2.36e-07 & 1.26e-10 & 8.62e-11 & 9.23e-10 \\
        & 10 & 1.14e-14 & 1.55e-14 & 1.66e-06 & 1.04e-08 & 1.71e-08 & 2.57e-15 & 1.43e-14 & 9.18e-13 \\
        & 20 & 3.74e-17 & 5.15e-22 & 2.93e-08 & 2.35e-09 & 4.31e-09 & 1.64e-14 & 4.27e-14 & 5.19e-12 \\
        & 40 & 1.09e-18 & 9.27e-24 & 1.32e-13 & 2.74e-13 & 1.87e-13 & 2.06e-18 & 3.70e-17 & 1.18e-15 \\ \hline
        \multirow{4}{5em}{Augmented}
        & 5 & 8.31e-10 & 9.36e-09 & 3.17e-06 & 6.57e-07 & 9.85e-08 & 1.04e-09 & 7.37e-11 & 3.30e-10 \\
        & 10 & 2.87e-15 & 1.55e-14 & 7.89e-06 & 2.95e-08 & 1.31e-08 & 4.85e-15 & 2.00e-14 & 5.14e-13 \\
        & 20 & 1.85e-18 & 2.92e-22 & 8.90e-08 & 3.80e-09 & 2.77e-09 & 4.48e-14 & 4.52e-14 & 2.27e-11 \\
        & 40 & 8.55e-19 & 2.88e-24 & 7.58e-13 & 4.44e-13 & 1.89e-13 & 2.90e-17 & 4.46e-17 & 1.19e-15 \\ \hline
        \multirow{4}{5em}{Haantjes}
        & 5 & 1.10e-09 & 1.56e-08 & 4.98e-06 & 2.23e-06 & 2.28e-07 & 1.41e-09 & 1.10e-10 & 1.03e-09 \\
        & 10 & 1.43e-14 & 1.68e-14 & 8.96e-07 & 1.49e-09 & 1.31e-08 & 5.01e-15 & 1.33e-14 & 2.56e-13 \\
        & 20 & 7.91e-19 & 2.27e-22 & 5.36e-08 & 1.58e-09 & 2.81e-09 & 3.27e-14 & 9.37e-14 & 8.55e-12 \\
        & 40 & 9.65e-19 & 3.61e-24 & 7.06e-12 & 5.19e-13 & 2.80e-13 & 4.43e-18 & 2.84e-17 & 3.17e-15 \\ \hline
        \multirow{4}{5em}{BFC}
        & 5 & 1.09e-09 & 9.42e-09 & 7.83e-07 & 6.44e-07 & 1.47e-07 & 1.49e-09 & 1.47e-10 & 4.30e-10 \\
        & 10 & 1.01e-14 & 1.29e-14 & 2.85e-06 & 1.13e-06 & 1.72e-07 & 1.07e-10 & 1.65e-10 & 4.81e-09 \\
        & 20 & 3.30e-17 & 1.86e-22 & 2.56e-08 & 2.29e-08 & 1.12e-08 & 3.87e-11 & 1.60e-11 & 1.06e-10 \\
        & 40 & 1.10e-18 & 8.71e-24 & 1.88e-13 & 2.40e-13 & 1.11e-13 & 3.98e-16 & 6.13e-17 & 3.86e-16 \\ \hline
        \end{tabular}
        \end{adjustbox}
    \caption{Minimum nonzero values of respective powers of normalized augmented adjacency matrices of datasets after SDRF-based rewiring using respective curvature types.}
    \label{table:powers}
\end{table}

\begin{figure}
\centering
\begin{subfigure}[t]{0.48\textwidth}
\centering
\includegraphics[width=0.885\linewidth]{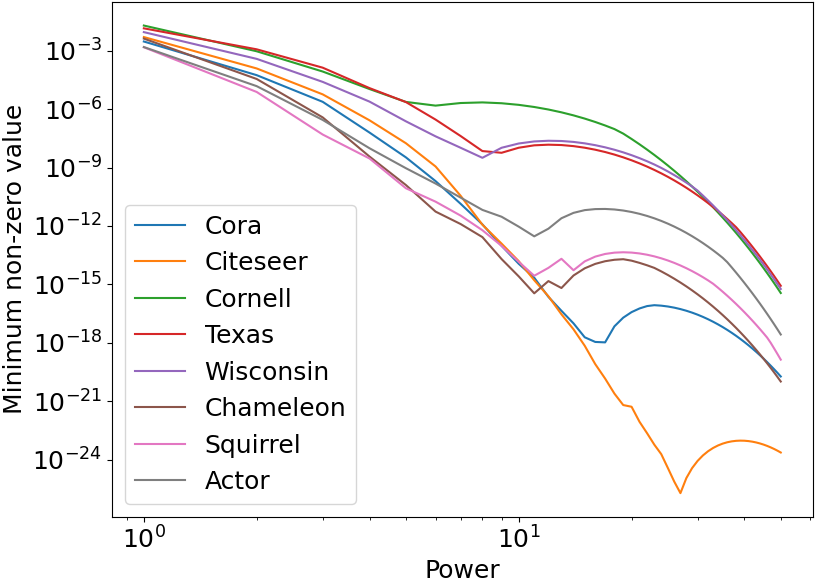}
\caption{1D}
\label{fig:adj-1d}
\end{subfigure}
\begin{subfigure}[t]{0.48\textwidth}
\centering
\includegraphics[width=0.9\linewidth]{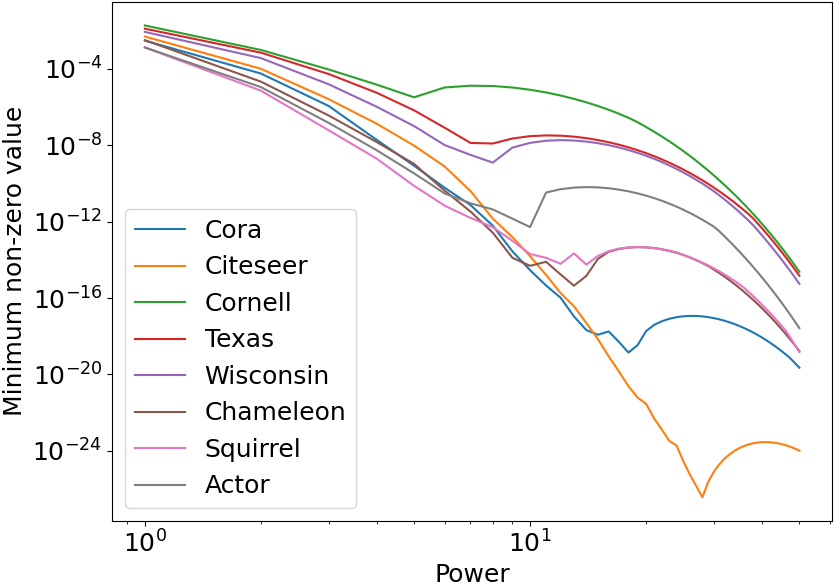}
\caption{Augmented}
\label{fig:adj-augmented}
\end{subfigure}

\br

\begin{subfigure}[t]{0.48\textwidth}
\centering
\includegraphics[width=0.875\linewidth]{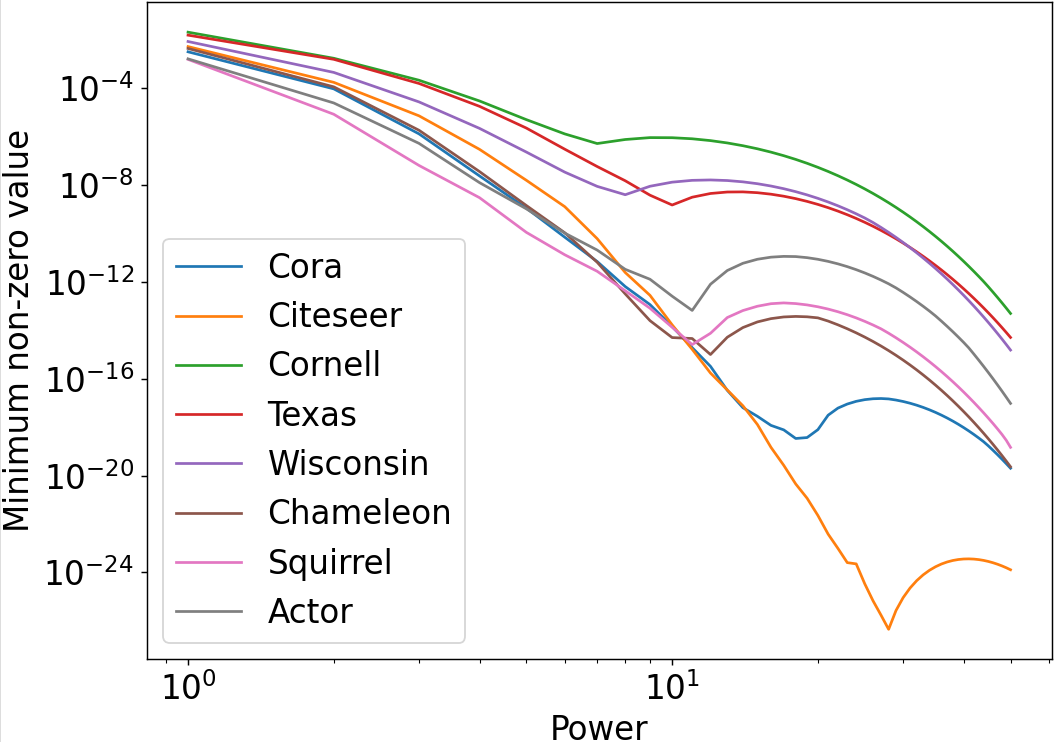}
\caption{Haantjes}
\label{fig:adj-haantjes}
\end{subfigure}
\begin{subfigure}[t]{0.48\textwidth}
\centering
\includegraphics[width=0.9\linewidth]{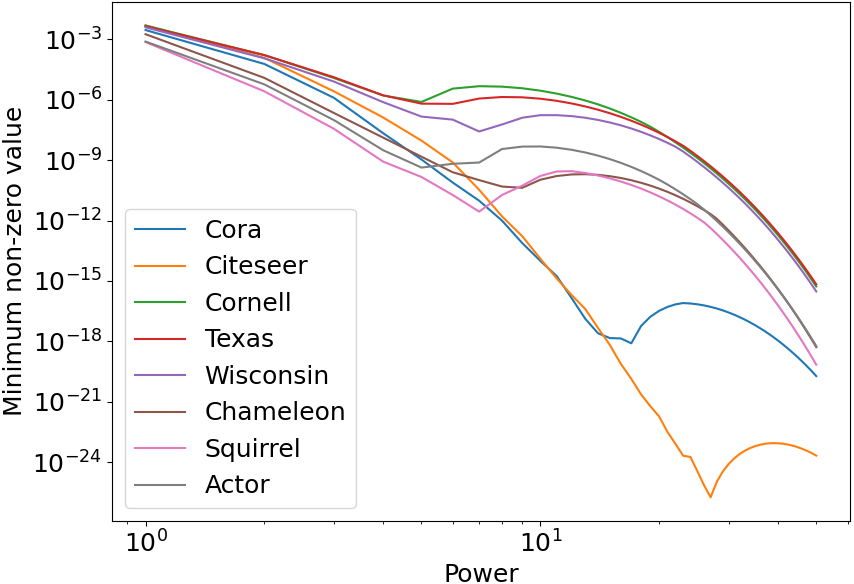}
\caption{BFC}
\label{fig:adj-bfc}
\end{subfigure}
\caption{Log-log plots of minimum non-zero values of powers of normalized augmented adjacency matrices after rewiring using different discrete curvature types.}
\label{fig:powers-curvatures}
\end{figure}

Figure~\ref{fig:powers-curvatures} represents analogous plots to that of Figure~\ref{fig:powers}, but for datasets after rewiring using each curvature type. These are graphical representations of the results in Table~\ref{table:powers}.

The rewiring instances chosen for this experiment are the ones with the best validation accuracy used for Table~\ref{table:best-rewirings}. For the datasets that were not evaluated (Chameleon, Squirrel, Actor), rewiring instances from the second run of experiment from Table~\ref{table:accuracies} were selected. The data presented in Table~\ref{table:powers} is rounded to $2$ decimal places. The matrix powers reported for each curvature type are chosen to be $5$, $10$, $20$ and $40$.

From Table~\ref{table:powers} and the comparison of Figure~\ref{fig:powers} to Figure~\ref{fig:powers-curvatures}, we see that rewiring successfully decreases the decay of the Jacobian bounds. The minimum nonzero values of powers of normalized augmented adjacency matrices are lower by many degrees of magnitude for no rewiring than for SDRF-based rewiring using any discrete curvature. The only two exceptions are Cora and Citeseer datasets, for which the values for respective powers are similar with and without rewiring; see Table~\ref{table:powers}. This also confirms that these two datasets have low bottleneckness and are not particularly prone to over-squashing.

%Moreover, it can be seen that the four plots in Figure~\ref{fig:powers-curvatures} are very similar to each other, both in terms of shapes of curves and scale, and the entries in Table~\ref{table:powers} are similar for corresponding powers and datasets for each curvature type that is not None.

\subsection{Computational Runtime}

We now assess computation runtime of the SDRF algorithm for graph rewiring based on each curvature type.  We measure the runtime for one rewiring process per curvature type and per dataset; the measurements are given in Table~\ref{table:times}.

\begin{table}
    \centering
    \begin{adjustbox}{max width=\textwidth}
    \centering
    \begin{tabular}{|l|l l l l l l|}
    \hline
        ~ & Cora & Citeseer & Pubmed & Cornell & Texas & Wisconsin \\ \hline
        1D & 5.86 & 5.31 & 53.57 & 0.34 & 0.41 & 0.61 \\
        Augmented & 6.16 & 5.48 & 107.55 & 0.19 & 0.21 & 0.49\\
        Haantjes & 2.88 & 4.27 & 39.65 & 0.13 & 0.13 & 0.14 \\ 
        BFC & 27.64 & 12.26 & OOM & 34.95 & 21.2 & 21.24 \\ \hline
    \end{tabular}
    \end{adjustbox}
    
    \br
    
    \centering
    \begin{adjustbox}{max width=\textwidth}
    \centering
    \begin{tabular}{|l|l l l l l l |}
    \hline
        ~ & Chameleon & Squirrel & Actor & Computers & Photo & Coauthor CS \\ \hline
        1D & 86.51 & 900 & 418.06 & 4369.07 & 853.52 & 47.15 \\ 
        Augmented & 251.25 & 901.71 & 872.78 & 10504.34 & 2262.72 & 88.10 \\ 
        Haantjes & 53.58 & 531.31 & 105.36 & 462.14 & 139.97 & 30.12\\ 
        BFC & 1627.61 & 2006.78 & 5121.31 & 6431.32 & 1274.44 & OOM \\ \hline
    \end{tabular}
    \end{adjustbox}
    
    \caption{Computation times of the SDRF rewiring given in seconds.}
    \label{table:times}
\end{table}

%The computation time has been recorded only for one instance of computation for each dataset and curvature type. This is because it would take an unreasonable amount of time to run the curvature computation e.g. $100$ times for each dataset and curvature type. This makes the results less robust, especially the results with low computation time. The reason is that the computation could take longer than expected for some datasets and curvature types due to unrelated reasons, like CPU or GPU occupancy with other processes. However, this is not an issue, as the short computation times are insignificant. The comparison between the long computation times is the one of interest, because it shows the difference in complexity of computation \textbf{at scale}. For these long computations, the random factors of CPU and GPU occupancy are not significant, and the relations between the computation times are reliable.

The runtimes here are reported for only one instance for each dataset and each curvature type, in order to avoid influences of spurious computational issues such as CPU and GPU occupancy with other processes which would become much more significant with repeated iterations.  Here, the interest is rather in the comparison between longer computation times which shows the difference in computational complexity at scale.

From these results, we see that all of the classical discrete curvatures studied have a significantly shorter computation time than the BFC. The slowest among the three classical curvatures is the augmented Forman curvature, at scale. This is expected, as it essentially needs to do the same calculations as 1D Forman and Haantjes curvatures combined (computation of degrees of endpoints and adjacent triangles for each edge).

For the Computers and Photo datasets, however, the computation of the augmented Forman curvature took longer than the computation of BFC. This suggests that for some types of graphs, possibly for bigger or more dense graphs (notice from Table~\ref{table:datasets} that the edges to nodes ratio is very high for these two datasets), the BFC computation can outperform the augmented Forman curvature computation in terms of computation time.  Nevertheless, the 1D Forman and Haantjes curvatures are still quicker to compute.

%%%%%%%%%%%%%%%%%%%%%%%%%%%%%%%%%%%%%%%%%%%%%%%%%%%%%%%%%%%%%%%%%%%%

\section{Discussion}
\label{sec:end}

In this paper, we systematically and comprehensively studied the role of various classical and novel discrete curvatures in mitigating the over-squashing problem in training GNNs.  Specifically, following the work of \cite{bottleneck-bronstein}, we adapted discretizations of Ricci curvature and Ricci flow, which can be viewed as the smooth, manifold-valued analogues of important characteristics on networks relevant in the information over-squashing problem---namely, information flow on a network and bottleneckness of a network, respectively.  In \cite{bottleneck-bronstein}---considered to be the current state-of-the-art in mitigating the over-squashing problem in GNN training, classified among the top 1.5\% of submissions in the 2022 International Conference on Learning Representations (ICLR) with Honorable Mention---the BFC was proposed as a discrete notion of Ricci curvature, while the SDRF algorithm was proposed as a discrete notion of Ricci flow.  In our work, we tested a wide range of classical discrete curvatures against the BFC in the implementation of the SDRF algorithm.  We found that more classical curvatures were able to achieve performance of the same order as the BFC in training accuracy and, at times, outperformed the BFC.  Moreover, they far outperformed it in computational runtime.  From this systematic study, we find that the impact of the contribution by \cite{bottleneck-bronstein} lies in the SDRF algorithm, rather than the BFC.  We also conclude that almost any of the more classical discrete curvatures may be used over the BFC together with the SDRF algorithm in favor of the more efficient computational runtimes, which is an important consideration when studying very large networks.

Directions of future study include exploring the performance of classical discrete curvatures taking into account directedness of the graphs in the SDRF and other rewiring methods.  Also, alternative discrete geometric approaches to mitigating the over-squashing problem that do not involve rewiring may be explored, in the spirit of the CGNN \citep{cgnn}.  Here, other computational notions of geometry for networks may also be investigated, such as those arising from topological data analysis, where persistent homology concurrently captures the topology of data as well as its integral geometry.  Such an approach would be particularly interesting when the goal is to preserve the topology of a graph, as the CGNN does.

%%%%%%%%%%%%%%%%%%%%%%%%%%%%%%%%%%%%%%%%%%%%%%%%%%%%%%%%%%%%%%%%%%%%

\newpage
\bibliographystyle{chicago}
\bibliography{oversquashing_ref.bib}

\end{document}